\PassOptionsToPackage{table}{xcolor}

\documentclass[runningheads]{llncs}

\usepackage{eccv}

\usepackage{eccvabbrv}

\usepackage{graphicx}
\usepackage{amsmath}
\usepackage{amssymb}
\usepackage{booktabs}
\usepackage{comment}

\usepackage{times}
\usepackage{epsfig}
\usepackage{float}
\usepackage{multirow}
\usepackage{amsfonts}
\usepackage{graphics}
\usepackage{witharrows}
\usepackage{arydshln}
\usepackage{subcaption} %

\usepackage[accsupp]{axessibility}  %

\usepackage[pagebackref,breaklinks,colorlinks]{hyperref}

\usepackage{orcidlink}

\usepackage[capitalize]{cleveref}
\crefname{section}{Sec.}{Secs.}
\Crefname{section}{Section}{Sections}
\Crefname{table}{Table}{Tables}
\crefname{table}{Tab.}{Tabs.}

\newcommand{\fref}[1]{Figure \ref{#1}}
\newcommand{\tref}[1]{Table \ref{#1}}

\newcommand{\eref}[1]{Equation \ref{#1}}

\begin{document}

\title{{\color{red}Mod}ality {\color{red}Tr}anslation for Object Detection Adaptation Without Forgetting Prior Knowledge} 

\titlerunning{ModTr for Object Detection}

\author{Heitor Rapela Medeiros\thanks{Email: heitor.rapela-medeiros.1@ens.etsmtl.ca}, Masih Aminbeidokhti, \\ Fidel Guerrero Pena, David Latortue, \\ Eric Granger, and Marco Pedersoli}

\institute{LIVIA, Dept. of Systems Engineering. ETS Montreal, Canada}

\authorrunning{Medeiros et al.}

\maketitle

\begin{abstract}
A common practice in deep learning involves training large neural networks on massive datasets to achieve high accuracy across various domains and tasks. While this approach works well in many application areas, it often fails drastically when processing data from a new modality with a significant distribution shift from the data used to pre-train the model. This paper focuses on adapting a large object detection model trained on RGB images to new data extracted from IR images with a substantial modality shift. We propose Modality Translator (ModTr) as an alternative to the common approach of fine-tuning a large model to the new modality. ModTr adapts the IR input image with a small transformation network trained to directly minimize the detection loss. The original RGB model can then work on the translated inputs without any further changes or fine-tuning to its parameters. Experimental results on translating from IR to RGB images on two well-known datasets show that our simple approach provides detectors that perform comparably or better than standard fine-tuning, without forgetting the knowledge of the original model. This opens the door to a more flexible and efficient service-based detection pipeline, where a unique and unaltered server, such as an RGB detector, runs constantly while being queried by different modalities, such as IR with the corresponding translations model. Our code is available at: \url{https://github.com/heitorrapela/ModTr}.

\end{abstract}

\begin{figure}
\centering
\begin{subfigure}[t]{0.24\textwidth}
    \caption{RGB - GT}
    \makebox[0pt][r]{\makebox[15pt]{\raisebox{30pt}{\rotatebox[origin=c]{90}{LLVIP}}}}%
    \includegraphics[width=\textwidth]
    {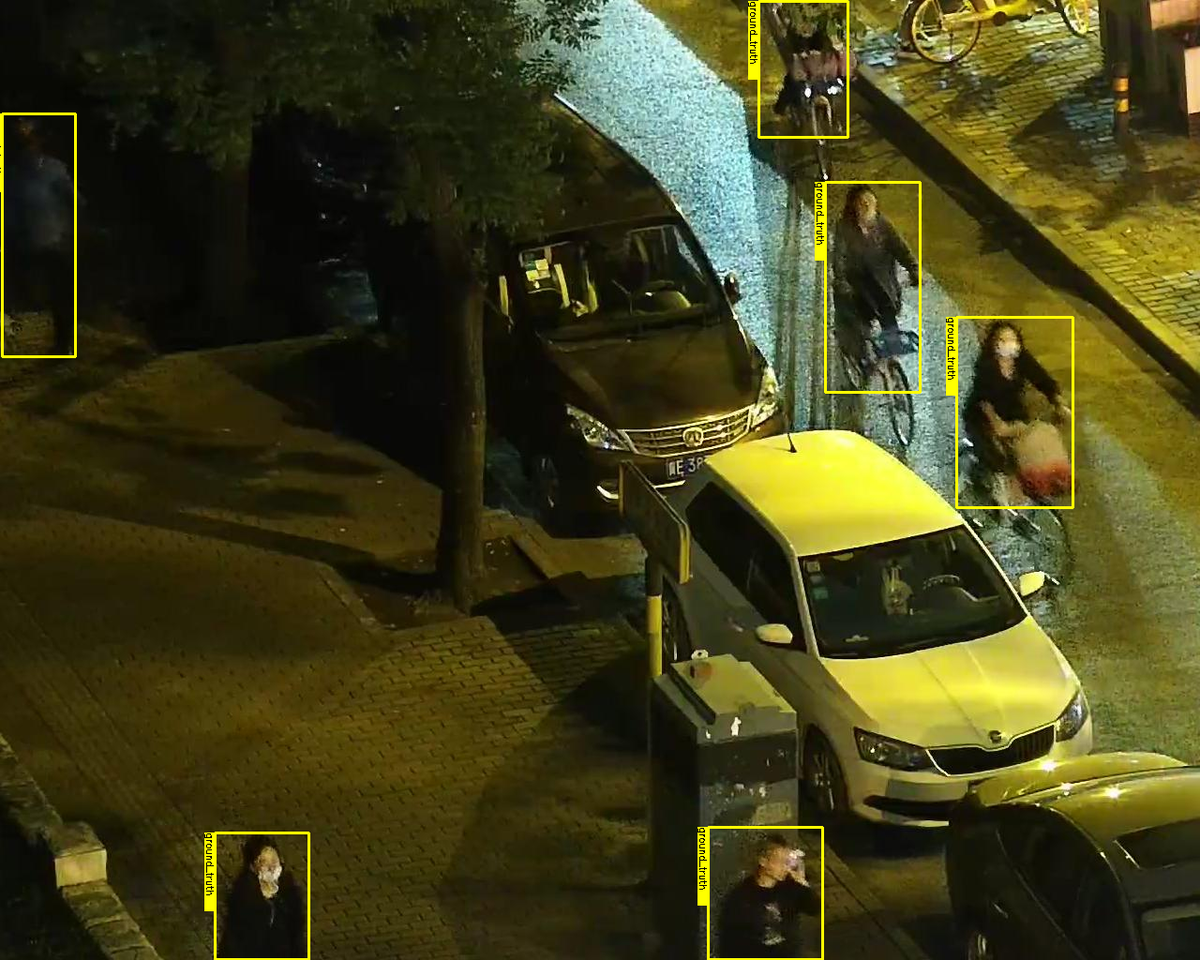}
    \makebox[0pt][r]{\makebox[15pt]{\raisebox{30pt}{\rotatebox[origin=c]{90}{FLIR}}}}%
    \includegraphics[width=\textwidth]
    {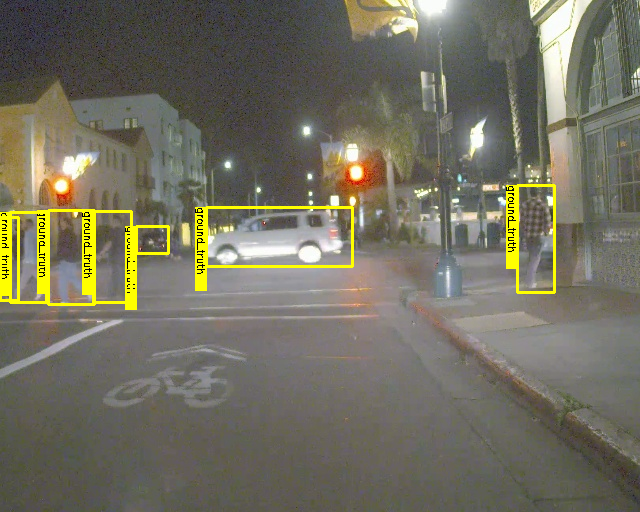}
\end{subfigure}
\begin{subfigure}[t]{0.24\textwidth}
    \caption{IR - FastCUT~\cite{park2020contrastive}}
    \includegraphics[width=\textwidth]  
    {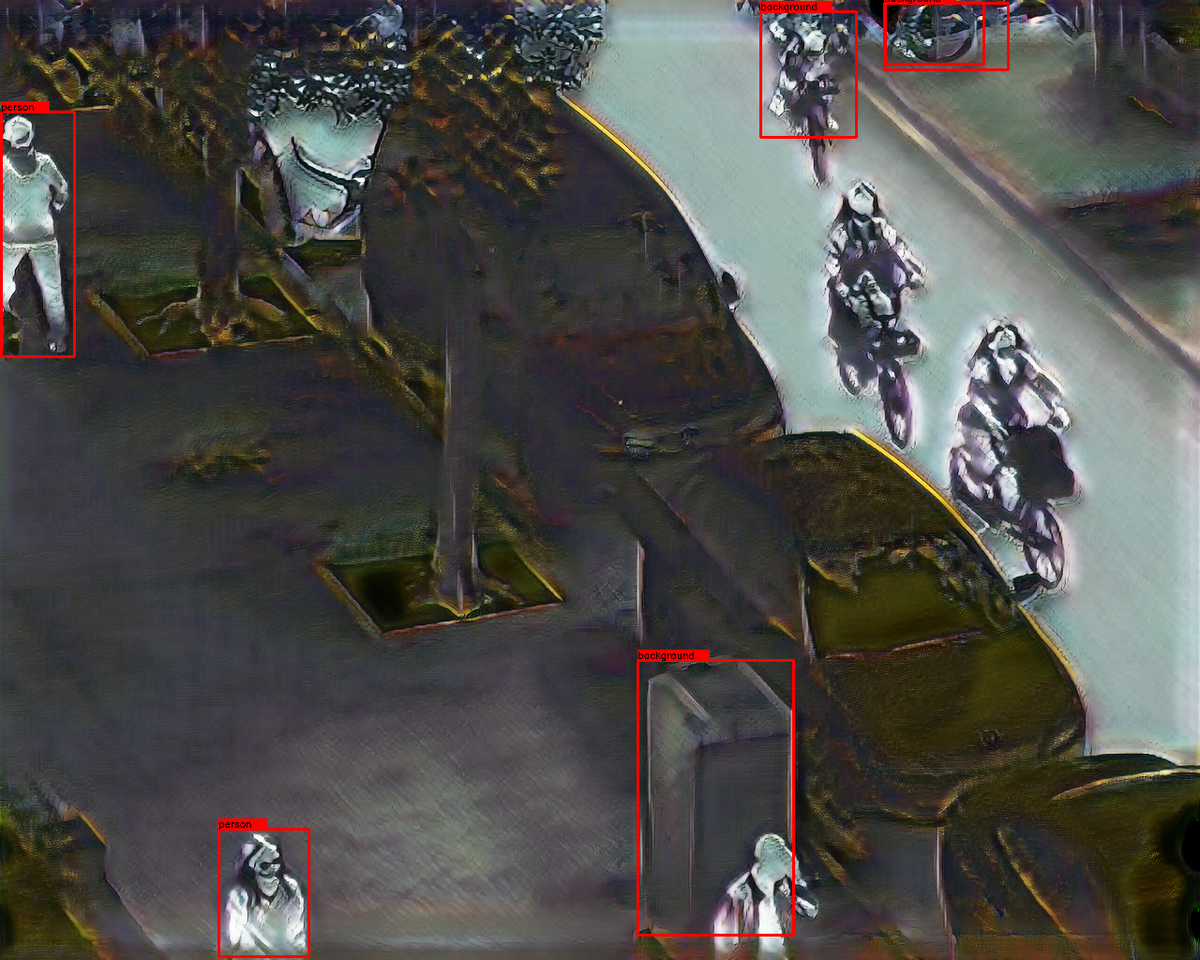}
    \includegraphics[width=\textwidth]
    {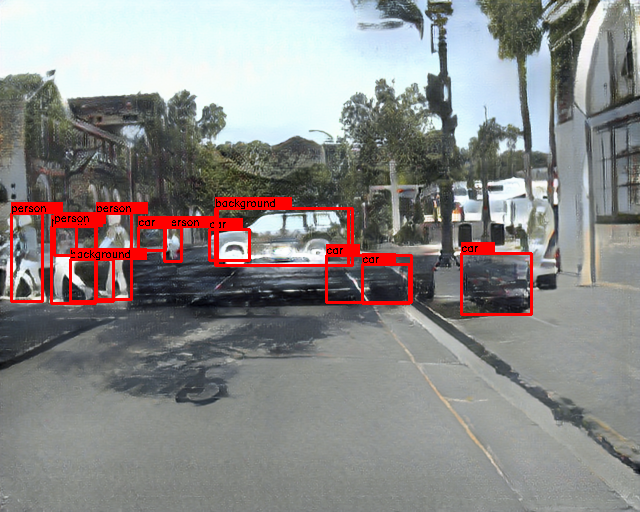}
\end{subfigure}
\begin{subfigure}[t]{0.24\textwidth}
    \caption{IR - Fine-tuning}
    \includegraphics[width=\textwidth]  
    {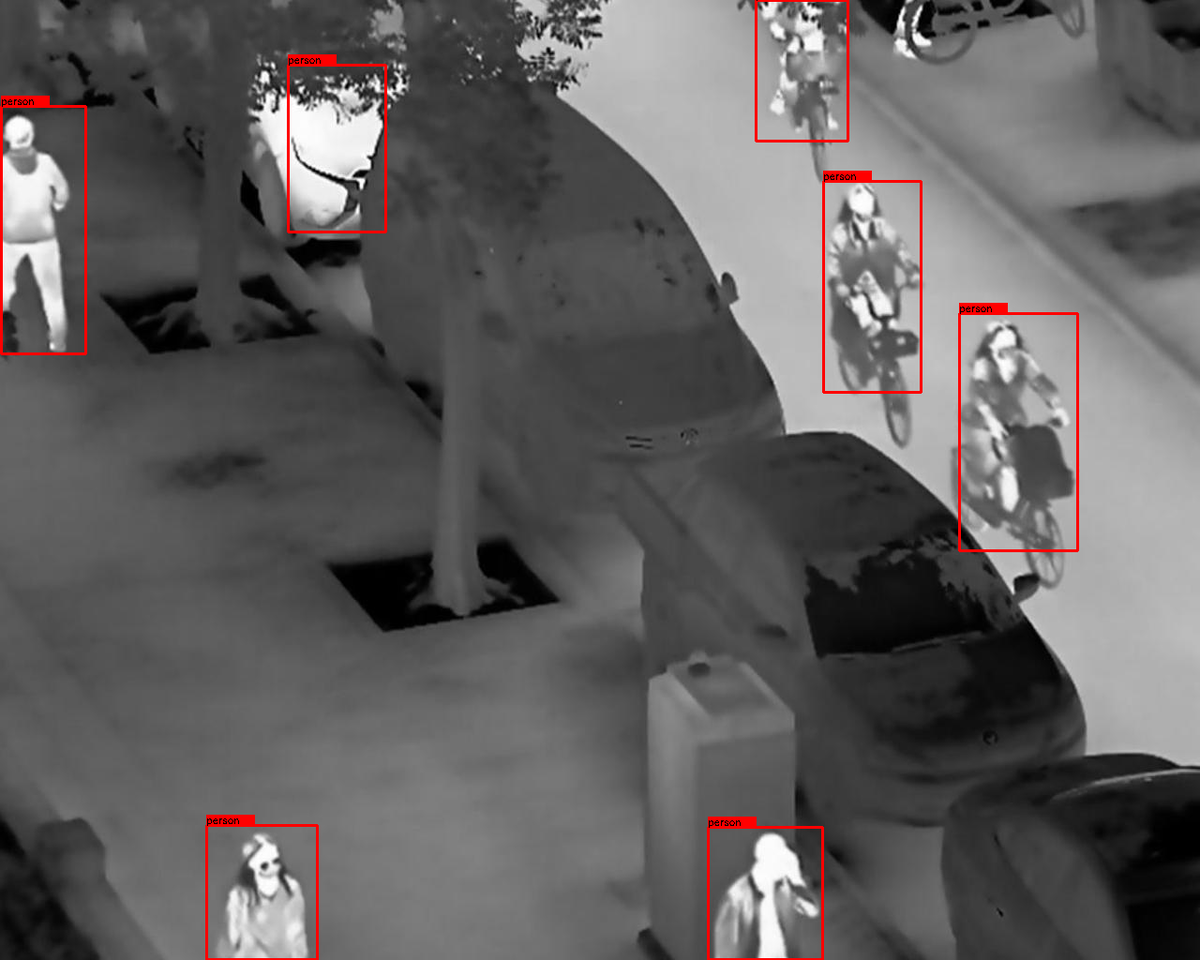}
    \includegraphics[width=\textwidth]
    {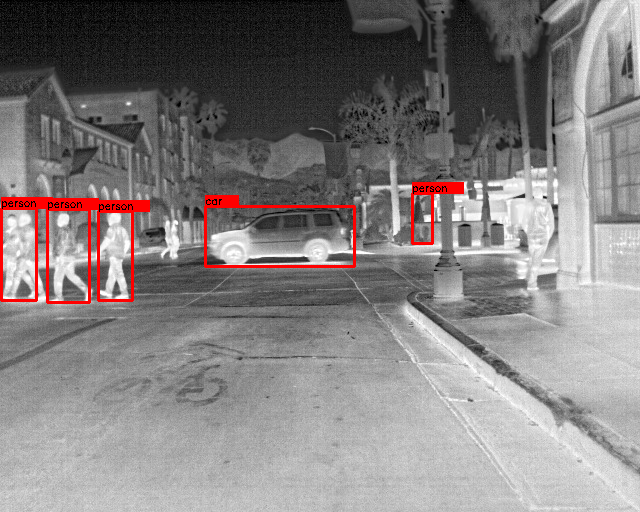}
\end{subfigure}
\begin{subfigure}[t]{0.24\textwidth}
    \caption{IR - ModTr (Ours)}
    \includegraphics[width=\textwidth]  
    {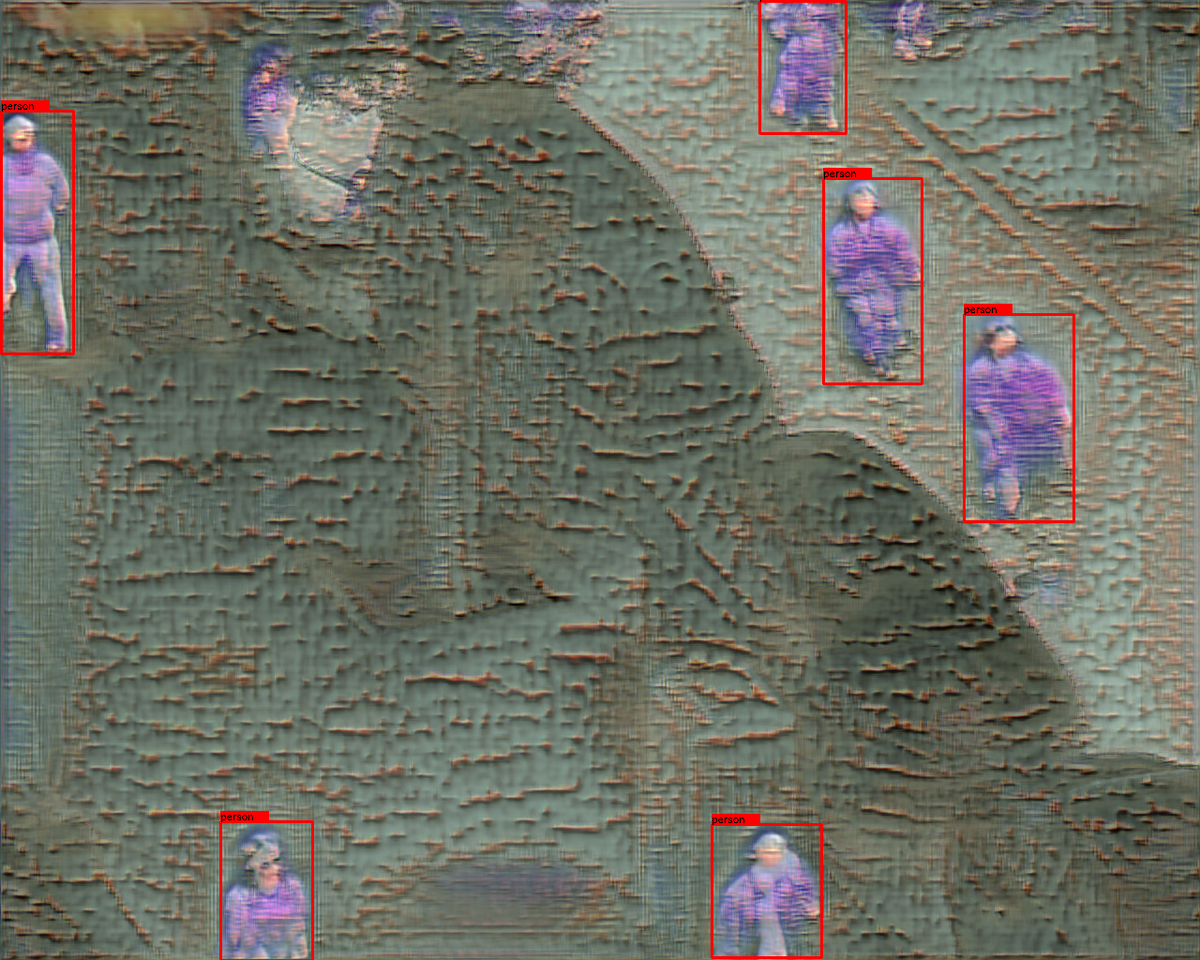}
    \includegraphics[width=\textwidth]
    {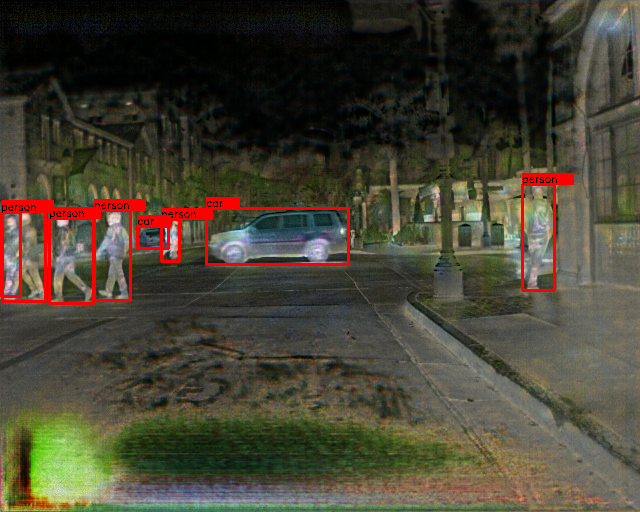}
\end{subfigure}
\caption{Bounding box predictions over different adaptations of the RGB detector (Faster R-CNN) for IR images on two benchmarks: LLVIP and FLIR. Yellow and red boxes show the ground truth and predicted detections, respectively. In a) we see the RGB data. In b) FastCUT is an unsupervised image translation approach that takes as input infrared images (IR) and produces pseudo-RGB images. It does not focus on detection and requires both modalities for training. In c) we have fine-tuning, which is the standard approach to adapting the detector to the new modality. It requires only IR data but forgets the original knowledge of the original RGB detector. Finally, in d) is the ModTr, which focuses the translation on detection, requires only IR data and does not forget the original knowledge so that it can be reused for other tasks. Bounding box predictions for other detectors are provided in the supplementary material.}
\label{fig:detections_results}
\end{figure}

\section{Introduction} \label{sec:intro}

Powerful pre-trained models have become essential in the field of computer vision, particularly in object detection (OD) tasks~\cite{minderer2022simple, minderer2024scaling}. These OD models are typically pre-trained on extensive natural-image RGB datasets, such as COCO~\cite{lin2014microsoft}. Moreover, the knowledge encoded by these models can be leveraged for various tasks in a zero-shot way or with additional fine-tuning for downstream tasks~\cite{vasconcelos2022proper}. However, adding new modalities to these models, such as infrared (IR), without losing the intrinsic knowledge of the detector remains a challenge~\cite{medeiros2024hallucidet}. 

These additional modalities, though not as common as RGB images, are still important in various tasks, like surveillance~\cite{chen2019distributed, dubail2022privacy}, autonomous driving~\cite{stilgoe2018machine, natan2022end}, and robotics~\cite{pierson2017deep, jing2017comparison}, which strive to achieve robust performance in real-world environments, where capture conditions change, such as different illumination conditions~\cite{bustos2023systematic}. The dominant way to adapt pre-trained detectors to these novel conditions is by fine-tuning the model~\cite{medeiros2024hallucidet}. However, fine-tuning often results in catastrophic forgetting and can destroy the intrinsic knowledge of the detector~\cite{kirkpatrick2017overcoming}. Ideally, we would like to adapt the detector to new modalities without changing the original model. This is most useful for server-side applications, where a single model runs uninterrupted and can be queried by different inputs, ideally on different modalities. The main challenge lies in the significant distribution shift introduced by the new modality. This shift occurs because the pre-trained knowledge, such as the visual information in RGB images, differs markedly from the thermal data in IR images. This shift can degrade model performance when applied directly as input to the model, since the features learned from one modality may not be relevant or present in another. This can ultimately impact the resulting OD performance~\cite{wang2022improving}.

Image translation methods~\cite{pang2021imagetoimage, park2020contrastive} have emerged as powerful tools to overcome the downsides of fine-tuning and narrowing the gap between source and target modalities~\cite{hsu2020progressive}. These methods do not directly work on the weight space of the original detector but rather adapt the input values to reduce the discrepancy between the source and target modalities. However, such methods often require access to source data or some statistics about it during training. Furthermore, their primary focus is on image reconstruction quality rather than the final OD task, which can cause a significant drop in performance. For instance, ~\fref{fig:detections_results} shows different ways to adapt the RGB detector (see the caption for more details).

\begin{figure}[!t]
        \centering
        \includegraphics[scale=0.37]{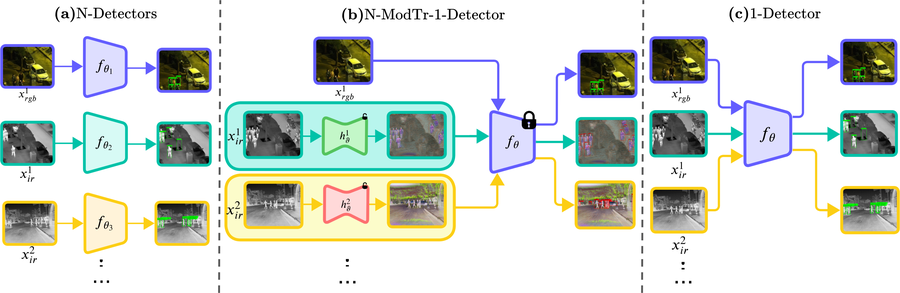}
    \caption{Different approaches to deal with multiple modalities and/or domains. (a) The simplest approach is to use a different detector adapted to each modality. This can lead to a high level of accuracy but requires storing several models in memory. (b) Our proposed solution uses a single pre-trained model normally trained on the more abundant data (RGB) and then adapts the input through our ModTr model. (c) A single detector is jointly trained on all modalities. This allows using of a single model but requires access to all modalities jointly, which is often impossible, especially when dealing with large pre-trained models.}
    \label{fig:different_methods}
\end{figure}

Our work aims to improve the image translation paradigm while addressing its limitations. Our proposed approach, Modality Translation for OD (ModTr), incorporates the detector's knowledge into the translation module by training directly for the final detection task. Unlike traditional image translation methods, ModTr does not require any source data. It is a conceptually simple approach that can be easily integrated with any detector, be it a one-stage or two-stage detector. A notable application of ModTr is using a pre-trained RGB detector as a server that incorporates different ModTr blocks as input translators for new modalities such as IR. This new detector generates the desired output with performance comparable to full fine-tuning without losing the original knowledge of the pre-trained model. In \cref{fig:different_methods}, we present several options for integrating IR modalities into an RGB system. \cref{fig:different_methods}a illustrates the N-Detectors approach, where each detector is trained for a specific case. This method effectively demands more memory and forgets previously learned information. \cref{fig:different_methods}c shows a single detector trained on combined modalities. This method does not incur additional memory, yet it requires simultaneous access to all modalities, which may not always be feasible. \cref{fig:different_methods}b illustrates our proposed approach, which involves training a specialized translator for each condition without altering the parameters of the original detector. The N-ModTr-1-Detector strikes a balance between the previous methods, addressing their shortcomings by requiring only a single detector. Importantly, it retains the original pre-training knowledge, as it leaves the detector unchanged. In this work, we focus on the effectiveness of our approach for the IR modality, commonly used in surveillance and robotics, and the incremental modality detector server-based application, which is crucial for many settings that require uninterrupted detection predictions.

\noindent \textbf{Our main contributions can be summarized as follows.} \\
\noindent \textbf{(1)}
We present ModTr, a method for adapting pre-trained ODs from large RGB datasets to new scarce modalities like IR, without requiring access to any source dataset, by translating the input signal.
\newline
\noindent \textbf{(2)}
In contrast to standard fine-tuning, our approach does not modify the original detector weights. This allows the detector to retain the knowledge of the source data while adapting to a new modality. As a result, a single model can be used to handle multiple modalities across various translators. For instance, the same model can be used to process RGB during the daytime and IR at nighttime.
\newline
\noindent \textbf{(3)} 
An extensive empirical evaluation of ModTr in several scenarios, showcasing its advantages and flexibility. In particular, with our different proposed fusion strategies, ModTr achieves OD accuracy that is competitive when compared with image translation methods on two challenging RGB/IR datasets (LLVIP and FLIR).

\section{Related Work} \label{sec:related_work}

\noindent \textbf{(a) Object Detection.} OD is a computer vision task that aims to provide labels and localization for the objects in the image~\cite{zhang2021dive}. Two-stage detectors, exemplified by Faster R-CNN~\cite{ren2015faster}, generate regions of interest and then use a second classifier to confirm object presence within those regions. On the other hand, one-stage detectors streamline the detection process by eliminating the proposal generation stage, aiming for end-to-end training and real-time inference speeds. RetinaNet~\cite{lin2017focal} is a one-stage OD model that utilizes a focal loss function to address class imbalance during training. Also, models like FCOS~\cite{tian2019fcos} have emerged in this category, eliminating predefined anchor boxes to potentially enhance inference efficiency. The proposed work investigates these three traditional and powerful detectors: Faster R-CNN, RetinaNet, and FCOS. The choice of such detectors was due to the simplicity in implementation and integration among other methods, as well as a different range of pre-trained backbone weights, such as ResNet~\cite{he2016deep} and MobileNet~\cite{howard2017mobilenets}.

\noindent \textbf{(b) Image Translation.} Image translation is a pivotal task in computer vision, aiming to map images from a source domain to a target domain while preserving inherent content~\cite{pang2021imagetoimage}. The goal is to discover a transformation function such that the distribution of images in the translated domain is aligned with the distribution of images in the target domain. The commonly used approaches for image translation are based on variational autoencoders (VAEs)~\cite{kingma2022autoencoding} and generative adversarial network (GANs)~\cite{goodfellow2014generative, pang2021imagetoimage}. Isola et al. developed the Pix2Pix~\cite{isola2017image}, a method that consists of a generator (based on U-Net) and a discriminator (based on GANs architecture) that work together to generate images based on input data and labels. Then, Zhu et al. proposed a method called CycleGAN~\cite{CycleGAN2017}, which is based on GANs, with the objective of unsupervised domain translation. Even though CycleGAN can produce quite visual results, it's hard to optimize due to the adversarial mechanism and memory footprint needed. In contrast, VAEs are easier to train than GANs but require more constraints in the optimization to produce images of good quality than GAN-based approaches. Recent advancements include diffusion models known for their high-quality image generation, although they may not inherently suit domain translation tasks. To enhance models such as CycleGAN, novel methods like Contrastive Unpaired Translation (CUT)~\cite{park2020contrastive} and FastCUT~\cite{park2020contrastive} have been introduced. CUT, in particular, accelerates the image translation process by maximizing mutual information between image patches, achieving competitive results quickly. In the context of RGB/IR modality, InfraGAN presents an image-level adaptation for RGB to IR conversion, prioritizing image quality~\cite{ozkanouglu2022infragan}. This approach is distinct in its focus on optimizing image quality losses. Moreover, Herrmann et al. have explored OD in RGB/IR modality by adapting IR images to RGB using traditional image preprocessing techniques, allowing the use of RGB object detectors without parameter modification~\cite{herrmann2018cnn}. Despite significant advances in image translation, these techniques do not specifically address OD tasks. In our previous work, we introduced HalluciDet~\cite{medeiros2024hallucidet}, which employs an image translation mechanism for OD. However, this approach requires prior access to the source RGB data from the same domain as the target for pre-training the detector.

\noindent \textbf{(c) Adapting Without Forgetting.} Catastrophic forgetting (CF) is the idea that a neural network tends to forget knowledge when sequentially trained on a different task and replaces it with knowledge tailored to the new objective~\cite{wang2023comprehensive}. CF can be harmful or beneficial. Researchers identified harmful learning as situations where retaining the original knowledge while adapting to a different task is necessary. In that case, it is imperative to mitigate the risk of CF. However, some CF can also be beneficial, for instance, to prevent privacy leakage from large pre-trained models, to enhance the generalization, or to remove noisy information from the originally, acquired knowledge that is negatively affecting the new tasks. In our case, knowledge-forgetting is harmful. There are different ways to address this issue including simple techniques like decreasing the learning rate~\cite{DBLP:journals/corr/abs-1801-06146}, use weight decay~\cite{CHELBA2006382, zhang2021revisiting} or mixout regularization~\cite{lee2020mixout} during fine-tuning or more complex approaches like Recall and learn~\cite{DBLP:journals/corr/abs-2004-12651}, Robust Information Fine-tuning~\cite{wortsman2022robust} or CoSDA~\cite{feng2023cosda}. Some adaptation methods use techniques based on replay of the source data or even using the weights of the initial model to keep some prior information~\cite{menezes2023continual}. Some of these works focus on adding continually different tasks in an incremental learning setting. However, these methods may still produce a loss of knowledge since the original parameters are not frozen. Furthermore, in adapting without forgetting, an adapter, which adopts a frozen pre-trained backbone to generate a representation followed by a different classifier for each downstream task~\cite{wang2023comprehensive}, can be seen as a powerful method to preserve knowledge. Even though our ModTr shares some similarities, we work in the input space to adapt to the new modalities, and address this incremental modality adaptation, optimizing the translation directly for the final OD task.

\section{Proposed Method} 
\label{sec:method}

\noindent \textbf{(a) Preliminary Definitions.} The training set for OD is denoted as $\mathcal{D}=\{(x, Y)\}$, where $x\in\mathbb{R}^{W\times H\times C}$ represents an image in the dataset, with dimensions $W\times H$ and $C$ channels. Subsequently, the OD model aims to identify $N$ regions of interest within these images, denoted as $Y=\{(b_i, c_i)\}_{i=1}^{N}$. The top-left corner coordinates and the width and height of the object define each region of interest $b_i$. Additionally, a classification label $c_i$ is assigned to each detected object, indicating its corresponding class within the dataset. In this study, the number of input channels for the detector is fixed at three, corresponding to RGB-like inputs. In terms of optimization, the primary goal of this task is to maximize detection accuracy, often measured using the average precision (AP) metric across all classes. An OD is formally represented as the mapping $f_{\theta}: \mathbb{R}^{W \times H \times C} \to \hat{Y}$, where $\theta$ denotes the parameter vector. To effectively train a detector, a differentiable surrogate for the AP metric, referred to as the detection cost function, $\mathcal{C}_{det}(\theta)$, is employed. The typical structure of such a cost function involves computing the average detection loss over dataset $\mathcal{D}$, denoted as $\mathcal{L}_{det}$, described as:

\begin{equation}
    \begin{split}
        \mathcal{C}_{det}(\theta)=\frac{1}{|\mathcal{D}|} \sum_{(x, Y) \in \mathcal{D}} \mathcal{L}_{det}(f_\theta(x), Y).
    \end{split} 
    \label{eq:detection_loss}
\end{equation}

\noindent \textbf{(b) Modality Translation Module.} Our approach primarily consists of an image-to-image translation network responsible for converting the input modality into an RGB-like space intelligible to the detector. These networks typically adopt an encoder-decoder structure to synthesize and reconstruct knowledge in a pixel-wise manner. While we employ U-Net~\cite{ronneberger2015u} as the translation network, with parameters $\vartheta$, in this work, our framework is general and not limited by the translation architecture. In general terms, this mapping is denoted as $h_{\vartheta}^d\colon \mathbb{R}^{W \times H \times C} \to \mathbb{R}^{W \times H \times 3}$, with a translation network assigned to each available input modality $d$. Unlike the detection network, the number of input channels varies depending on the modality, for instance, $C=1$ for IR and depth images. It's important to note that, being a pixel-level architecture, the output of such a network retains the spatial resolution of the input. However, the number of output channels is consistently fixed at three, corresponding to RGB-like images ($C=3$). 

Unlike other image-to-image translation approaches, we drive the process using the aforementioned detection cost (Equation~\eqref{eq:detection_loss}). Thus, the underlying optimization problem is formulated as $\vartheta^* = \arg\min \mathcal{L}_{det}(\vartheta)$, incorporating the output of the composition $(f_\theta \circ h_\vartheta^d)(x)$ at the loss function level. To streamline the learning process, we utilize a residual learning strategy in which the function $h_\vartheta^d$ focuses on capturing the small variations in the input that are necessary to solve the task. This approach is similar to the one employed on diffusion models, which inspired our work. For the sake of simplicity, we separate the fusion step from the translation mapping in our notation, as various types of fusion are investigated. Consequently, the proposed image-to-image translation loss function is defined as:  

\begin{equation} 
\label{eq:modtr_loss}
\begin{split}
\mathcal{L}_{\text{ModTr}}(x, Y; \vartheta) & = \mathcal{L}_{det}(f_\theta\left(\Phi(h_\vartheta^d(x),x)\right), Y),
\end{split}
\end{equation}
\noindent where $\Phi(.,.)$ is a non-parametric fusion function. Note that the output of $h_\vartheta^d(x)$ is an RGB-like image, whereas $x$ may only consist of a single channel, depending on the input modality. We have chosen this definition to simplify the notation, but appropriate reshaping should be performed during implementation to ensure compatibility. 

In addition, note that, while a detection loss is employed to update the translation network, the weight vector $\theta$ remains constant. This constraint is consistent with the premise of this study, where a pre-trained detector is solely available on the server side and remains unaltered. An overview of the proposed approach can be seen in~\cref{fig:different_methods} b).

\noindent \textbf{(c) Fusion strategy.} As previously mentioned, we utilize a non-parametric fusion of the intermediate representation $h_\vartheta^d(x)$ and the original input $x$ to simplify the learning process of the translation network. In this context, we employ an element-wise product, also known as the Hadamard product, which is particularly interesting for attention mechanisms and has been explored previously for re-calibrating feature maps based on their importance~\cite{hu2018squeeze}. Although we investigated various fusion mechanisms, the element-wise product yielded the best results. For more details on different fusion strategies, please refer to supplementary materials.

\noindent \textbf{ModTr$_{\odot}$}: The Hadamard product-based fusion serves as a gating mechanism to filter or highlight information from the input image. In this approach, the output of the translation network acts as a weight map for the input, and they are fused using pixel-wise multiplication, $\odot$. Consequently, the translation network tends to highlight information from the input when the pixel value tends toward $1$ or discard it when it approaches $0$. Additionally, the output translation modality can be interpreted as an attention map, as described by the following Equation~\eqref{eq:modtr_product}:
\begin{equation} 
\label{eq:modtr_product}
\begin{split}
\mathcal{L}_{\text{ModTr}_{\odot}}(x, Y; \vartheta) & = \mathcal{L}_{det}(f_\theta\left(h_\vartheta^d(x)\odot x\right), Y).
\end{split}
\end{equation}

In our design choices, we opt to utilize these straightforward non-parametric functions to assist in optimization while maintaining low inference costs.

\setlength{\tabcolsep}{8pt}
\renewcommand{\arraystretch}{1.0}

\section{Results and Discussion}  \label{sec:experiments}

\subsection{Experimental Methodology}

\subsubsection{(a) Datasets:} \textbf{LLVIP:} LLVIP is a surveillance dataset composed of $30,976$ images, in which $24,050$ ($12,025$ IR and $12,025$ RGB paired images) are used for training and $6,926$ for testing ($3,463$ IR and $3,463$ RGB paired images) with only pedestrians annotated. \textbf{FLIR ALIGNED:} We used the sanitized and aligned paired sets provided by Zhang et al.~\cite{zhang2020multispectral}. It has $10,284$ images, that is $8,258$ for training ($4,129$ IRs and $4,129$ RGBs) and $2,026$ ($1,013$ IRs and $1,013$ RGBs) for test. FLIR images are captured from the perspective of a camera in the front of a car, with a resolution of $640$ by $512$. It contains the bicycles, dogs, cars, and people classes. It has been found that with FLIR, the "dog" objects are inadequate for training~\cite{cao2023multimodal}, thus we decided to remove them.

\subsubsection{(b) Implementation details:}
In our experiments, we randomly selected $80\%$ of the training set for training and the rest for validation. All results reported are on the test set. As starting pre-trained weights for the detectors, we used Torchvision models with COCO~\cite{lin2014microsoft} weights and for the U-Net translation network, we used PyTorch Segmentation Models~\cite{Iakubovskii:2019} and we changed the last layer for 3-channel (RGB-like) with a Sigmoid function, to be closer to an image with values between $0$ and $1$, to perform translation instead of traditional segmentation. For the translation network backbones, we explored our default ResNet$_{34}$, and for subsequent studies on reducing parameters, we dive into ResNet and MobileNet-family. All the code is available on GitHub for reproducibility in the experiments. To ensure fairness, we trained the detectors under the library version and the same experimental design, i.e., data order, augmentations, etc. Furthermore, we trained with PyTorch Lightning~\cite{Falcon_PyTorch_Lightning_2019} training framework, evaluated the APs with TorchMetrics~\cite{detlefsen2022torchmetrics}, and logged all experiments with WandB~\cite{wandb} logging tool. The different performance measures (e.g., APs) can be found in suppl. materials.

\subsection{Comparison with Translation Approaches}

In this section, ModTr is compared with different image-to-image translation methods employing different learning strategies. These include basic image processing strategy\cite{herrmann2018cnn}, reconstruction strategies such as CycleGAN~\cite{zhu2017unpaired}, CUT~\cite{park2020cut}, and FastCUT~\cite{park2020cut}, which employs a contrastive learning approach, as well as HalluciDet~\cite{medeiros2024hallucidet}, which utilizes a detection-based loss. As outlined in~\tref{tab:img2img_main_table}, we evaluated the methods based on their final detection performance across three commonly used detectors: FCOS, RetinaNet, and Faster R-CNN. The reported results are derived from the IR test set and are averaged over three different seeds, which helps mitigate the impact of randomness across runs and splits of the training and validation datasets.

For each method, we also consider its dependency on the prior knowledge data (RGB) and ground truth bounding boxes (bboxes) on the IR images. Methods that rely on reconstruction techniques do not require bbox annotations on IR images but cannot provide accurate translations for detection purposes. However, HalluciDet and ModTr require bbox annotations to adjust the input image in a discriminative manner. The main difference between HalluciDet and ModTr is the use of source images. HalluciDet requires RGB images for an initial fine-tuning of the model, while our approach can work without that fine-tuning by reusing the detector's zero-shot knowledge.

The proposed ModTr displays robustness across the three detectors and consistently exhibits improvement on two different datasets: LLVIP~\cite{jia2021llvip} and FLIR aligned~\cite{fa2018flir}. Note that each algorithm described in~\tref{tab:img2img_main_table} employs different training supervisions. For instance, CycleGAN employs an adversarial mechanism with both RGB and infrared modalities in an unpaired setting. Similarly, CUT and FastCUT operate with positive and negative patches in an unpaired setting. In contrast, HalluciDet doesn't require the presence of both modalities during training but employs a detection mechanism during training similar to ours. In our approach, we solely require examples from the target modality. In this section, we present the performance of our best approach $\text{ModTr}_{\odot}$. For additional results, refer to suppl. materials.

\begin{table}[h]
\caption{Detection performance (AP) of ModTr versus baseline image-to-image methods to translate the IR to RGB-like images, using three different detectors (FCOS, RetinaNet, and Faster R-CNN).
The methods were evaluated on IR test set of LLVIP and FLIR datasets. The RGB column indicates if the method required access to RGB images during training, and Box refers to the use of ground truth boxes during training.
}
\label{tab:img2img_main_table}
    \centering
    \resizebox{1.0\textwidth}{!}{%
    \begin{tabular}{lccccc}

        \toprule
        
        \multirow{2}{*}[-1em]{\textbf{Image translation}} & \multirow{2}{*}[-1em]{\textbf{RGB}} & \multirow{2}{*}[-1em]{\textbf{Box}} & \multicolumn{3}{c}{\textbf{Test Set IR (Dataset: LLVIP)}} \\
        \cmidrule(lr){4-6}
        \addlinespace[5pt]        
        {} & {} & {} & \multicolumn{1}{c}{\multirow{2}{*}[1em]{\textbf{FCOS}}} & \multicolumn{1}{c}{\multirow{2}{*}[1em]{\textbf{RetinaNet}}} & \multicolumn{1}{c}{\multirow{2}{*}[1em]{\textbf{Faster R-CNN}}} \\

        \midrule
        
        Histogram Equal.~\cite{herrmann2018cnn}  & {} & {} & 31.69 ± 0.00 & 33.16 ± 0.00 & 38.33 ± 0.02 \\
        \rowcolor[HTML]{EFEFEF}

        CycleGAN~\cite{zhu2017unpaired}& \checkmark & {} &  23.85 ± 0.76 & 23.34 ± 0.53 & 26.54 ± 1.20 \\

        CUT~\cite{park2020cut}& \checkmark & {} & 14.30 ± 2.25 & 13.12 ± 2.07 & 14.78 ± 1.82 \\
        
        \rowcolor[HTML]{EFEFEF}
        FastCUT~\cite{park2020cut}& \checkmark & {} &  19.39 ± 1.52 &  18.11 ± 0.79 & 22.91 ± 1.68 \\

        HalluciDet~\cite{medeiros2024hallucidet}& \checkmark& \checkmark & 28.00 ± 0.92 & 19.95 ± 2.01 & 57.78 ± 0.97 \\
        \midrule
        \rowcolor[HTML]{EFEFEF}
        ModTr$_{\odot}$ (ours) & {} & \checkmark &  \textbf{57.63 ± 0.66} & \textbf{54.83 ± 0.61} & \textbf{57.97 ± 0.85} \\

        \toprule
        
        \multirow{2}{*}[-1em]{\textbf{Image translation}} &  \multirow{2}{*}[-1em]{\textbf{RGB}} & \multirow{2}{*}[-1em]{\textbf{Box}} & \multicolumn{3}{c}{\textbf{Test Set IR (Dataset: FLIR)}} \\
        \cmidrule(lr){4-6}
        \addlinespace[5pt]        
        {} & {} & {} & \multicolumn{1}{c}{\multirow{2}{*}[1em]{\textbf{FCOS}}} & \multicolumn{1}{c}{\multirow{2}{*}[1em]{\textbf{RetinaNet}}} & \multicolumn{1}{c}{\multirow{2}{*}[1em]{\textbf{Faster R-CNN}}} \\

        \midrule
        Histogram Equal.~\cite{herrmann2018cnn} & {} & {} & 22.76 ± 0.00 & 23.06 ± 0.00 & 24.61 ± 0.01 \\
        \rowcolor[HTML]{EFEFEF}
        CycleGAN~\cite{zhu2017unpaired}&   \checkmark & {} & 23.92 ± 0.97 & 23.71 ± 0.70 & 26.85 ± 1.23 \\
        CUT~\cite{park2020cut}&  \checkmark & {} &  18.16 ± 0.75 & 17.84 ± 0.75 & 20.29 ± 0.48 \\
        \rowcolor[HTML]{EFEFEF}
        FastCUT~\cite{park2020cut}&  \checkmark & {} &   24.02 ± 2.37 & 22.00 ± 2.73 & 26.68 ± 2.59 \\
        HalluciDet~\cite{medeiros2024hallucidet}  & \checkmark & \checkmark & 23.74 ± 2.09 & 22.29 ± 0.45 & 29.91 ± 1.18 \\
        \midrule

        \rowcolor[HTML]{EFEFEF}
        ModTr$_{\odot}$ (ours) & {} & \checkmark &  \textbf{35.49 ± 0.94} & \textbf{34.27 ± 0.27} & \textbf{37.21 ± 0.46} \\
        
        \bottomrule
    \end{tabular}
    }

\end{table}

As reported in~\tref{tab:img2img_main_table}, the detection performance of ModTr over the LLVIP dataset exhibited significant improvements. Specifically, it surpassed HalluciDet, the second best, by more than $29.0$ AP with both FCOS and RetinaNet architectures, while obtaining comparable results with Faster R-CNN. Such disparity with the previous technique can be attributed to the loss of previous knowledge inherent in HalluciDet, which necessitates a pre-fine-tuning strategy on the source modality. 
Although the performance of the FLIR dataset also improved, the dataset's inherent challenges, such as changing the background from a moving car setup, make detection more difficult. Nonetheless, our proposal consistently enhances results, with improvements of more than $11$ AP for FCOS and RetinaNet, and over $7$ AP for Faster R-CNN. We also observed improvements on the AP$_{50}$ and AP$_{75}$. Because of the space constraint, we include these in supplementary materials.
These promising results indicate that our proposal can effectively translate images from the original IR modality to an RGB-like representation, sufficiently close to the source data to be usable by the detector.

\subsection{Translation vs. Fine-tuning}

In this section, we further show that the proposed approach can be trained jointly with both translation and detector, which preserves the detector's knowledge. Here, ModTr is compared to three baselines fully fine-tuning (FT), FT of the head and LoRA~\cite{hu2022lora}, and our best ModTr fusion strategy, as shown in ~\cref{tab:flir_tab2}. 

We conduct LoRA fine-tuning using two settings. In the first, we apply LoRA across all layers; in the second, only to the last layer of detectors. The latter results in superior performance, so we have adopted it as our default LoRA setting. The ~\cref{tab:flir_tab2} shows AP for the LLVIP and FLIR datasets, with a consistent trend across all detectors (FCOS, RetinaNet, and Faster R-CNN). Furthermore, in the case of the FLIR dataset, we observed enhancements of ModTr over the standard detector FT. As demonstrated, our approach surpasses standard fine-tuning while maintaining the detector's performance in the original modality. It is worth noting that our method also improves performance in terms of localization metrics such as AP$_{50}$ and AP$_{75}$ compared to fine-tuning alone, and we provide detailed results in the supplementary materials.

\begin{table}[h]
\caption{Detection performance (AP) of ModTr versus baseline fine-tuning (FT) of the detector, FT of the head and LoRA~\cite{hu2022lora}, using three different detectors (FCOS, RetinaNet, and Faster R-CNN. The methods were evaluated on IR test set of LLVIP and FLIR datasets. Results with "-" diverged from the optimization.}

\label{tab:flir_tab2}
    \centering
    \resizebox{0.8\textwidth}{!}{%
    \begin{tabular}{lccc}
        \toprule
        \multirow{2}{*}[-1em]{\textbf{Method}} &  \multicolumn{3}{c}{\textbf{Test Set IR (Dataset: LLVIP)}} \\
        \cmidrule(lr){2-4}
        \addlinespace[5pt]        
        {} & \multicolumn{1}{c}{\multirow{2}{*}[1em]{\textbf{FCOS}}} & \multicolumn{1}{c}{\multirow{2}{*}[1em]{\textbf{RetinaNet}}} & \multicolumn{1}{c}{\multirow{2}{*}[1em]{\textbf{Faster R-CNN}}} \\
        \midrule

        Fine-Tuning (FT) & 57.37 ± 2.19 & 53.79 ± 1.79 & \textbf{59.62 ± 1.23} \\

        FT Head &  49.11 ± 0.70   & 44.00 ± 0.28  & 59.33 ± 2.17 \\

        LoRA~\cite{hu2022lora}  &  47.72 ± 0.58 & - & 54.83 ± 1.30 \\
        
        \midrule
        
        \rowcolor[HTML]{EFEFEF}
        ModTr$_{\odot}$ (ours) & \textbf{57.63 ± 0.66} & \textbf{54.83 ± 0.61} & 57.97 ± 0.85 \\

        \toprule

      \multirow{2}{*}[-1em]{\textbf{Method}} &  \multicolumn{3}{c}{\textbf{Test Set IR (Dataset: FLIR)}} \\
        \cmidrule(lr){2-4}
        \addlinespace[5pt]        
        {} & \multicolumn{1}{c}{\multirow{2}{*}[1em]{\textbf{FCOS}}} & \multicolumn{1}{c}{\multirow{2}{*}[1em]{\textbf{RetinaNet}}} & \multicolumn{1}{c}{\multirow{2}{*}[1em]{\textbf{Faster R-CNN}}} \\

        \midrule

        Fine-Tuning (FT) & 27.97 ± 0.59 & 28.46 ± 0.50 & 30.93 ± 0.46 \\

        FT Head & 27.40 ± 0.12 & 26.78 ± 0.70 & 33.53 ± 0.36 \\

        LoRA~\cite{hu2022lora} &  -   & -  & 29.44 ± 0.61 \\
        
        \midrule
        
        \rowcolor[HTML]{EFEFEF}
        ModTr$_{\odot}$ (ours) &  \textbf{35.49 ± 0.94} & \textbf{34.27 ± 0.27} & \textbf{37.21 ± 0.46} \\

        \bottomrule
    \end{tabular}
    }
\end{table}

\subsection{Different Backbones for ModTr}

In this context, we evaluate ModTr and examine the trade-off between performance and parameter cost. It is widely recognized that increasing the number of parameters can enhance performance, but this relationship is not strictly linear. We demonstrated that models with fewer parameters can still achieve good performance; for example, MobileNet$_{v2}$, with fewer parameters than ResNet$_{18}$, sometimes outperformed it. This trade-off highlights the versatility of the model, which can be deployed with MobileNet-based architectures and utilized in low-cost devices. In \tref{tab:diff_backbones}, the default number of parameters is successfully reduced from $24.4$M (ResNet$_{34}$) to $6.6$M using MobileNet$_{v2}$ while maintaining similar performance. For instance, on LLVIP, MobileNet$_{v2}$ achieved a mean AP of $56.15$, comparable to $56.35$ AP$_{50}$ from ResNet$_{34}$ (others APs and detectors are reported in the supplementary material).

This approach opens up new possibilities, particularly in scenarios where using one translation network and one detector (e.g., one ModTr and one detector for RGB/IR) proves advantageous. This setup requires a total of $44.9$M parameters, compared to $83.6$M parameters, when employing two detectors—one for each modality (for example, for Faster R-CNN). Similar reductions in parameter costs were observed for FCOS (from $66.4$M to $36.3$M) and RetinaNet (from $68$M to $37.1$M) when using one detector for both modalities while preserving the knowledge of the previous modality and incorporating a new one. These numbers are based on MobileNet$_{v3s}$, which strikes a balance between performance and the number of parameters, making it suitable for memory-restricted systems. The complete evaluations for FCOS and RetinaNet are included in the supplementary material.

\newcolumntype{g}{>{\columncolor[HTML]{EFEFEF}}c}
\newcolumntype{q}{>{\columncolor[HTML]{EFEFEF}}l}

\begin{table}[h]

\caption{Detection performance (AP) of ModTr with different backbones for the translation networks with different numbers of parameters,  using three different detectors (FCOS, RetinaNet, and Faster R-CNN). The methods were evaluated on IR test set of LLVIP and FLIR datasets.}
\label{tab:diff_backbones}
    \centering    
    \resizebox{0.8\textwidth}{!}{%
    \begin{tabular}{qgg}

        \toprule
        \rowcolor{white}
        {} & \multicolumn{1}{c}{\textbf{Test Set IR (Dataset: LLVIP)}} & {}  \\
        
        \midrule
        \rowcolor{white}
        \multicolumn{1}{c}{\textbf{Method}}  & \textbf{Parameters} & \textbf{AP${}\uparrow$} \\
        \midrule
        \addlinespace[5pt]
        \rowcolor{white}

        \multicolumn{1}{c}{\textbf{Faster R-CNN}} & 41.8 M & {} \\
        \midrule 
        MobileNet$_{v3s}$ & + 3.1 M & 54.51 ± 0.28 \\
        \rowcolor{white}
        MobileNet$_{v2}$ & + 6.6 M & 56.15 ± 0.51  \\
        
        ResNet$_{18}$ & + 14.3 M & 55.53 ± 1.14 \\
        \rowcolor{white}
        ResNet$_{34}$ & + 24.4 M  & 56.35 ± 0.65 \\

        \midrule
        \rowcolor{white}
        {} & \multicolumn{1}{c}{\textbf{Test Set IR (Dataset: FLIR)}} & {}   \\

        \midrule

        \addlinespace[5pt]
        \rowcolor{white}
        \multicolumn{1}{c}{\textbf{Faster R-CNN}} & 41.8 M & {}  \\
        \midrule 
        MobileNet$_{v3s}$ & + 3.1 M & 32.06 ± 0.75 \\
        \rowcolor{white}
        MobileNet$_{v2}$ & + 6.6 M & 36.77 ± 0.67 \\
        ResNet$_{18}$ & + 14.3 M & 36.68 ± 0.22 \\
        \rowcolor{white}
        ResNet$_{34}$ & + 24.4 M   & 37.21 ± 0.46 \\

        \bottomrule
    \end{tabular}
    }
\end{table}

\begin{figure*}[!ht]
\centering
    \begin{tabular}{c}
  \toprule
    LLVIP Test Dataset \\
    \midrule 

    \makebox[0pt][r]{\makebox[10pt]{\raisebox{12pt}{\rotatebox[origin=c]{90}{\scriptsize GT}}}}
    \includegraphics[width=0.95\textwidth]{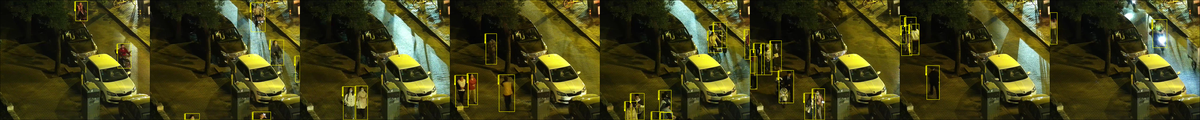} \\

    \makebox[0pt][r]{\makebox[10pt]{\raisebox{15pt}{\rotatebox[origin=c]{90}{\scriptsize FT}}}}
    \includegraphics[width=0.95\textwidth]{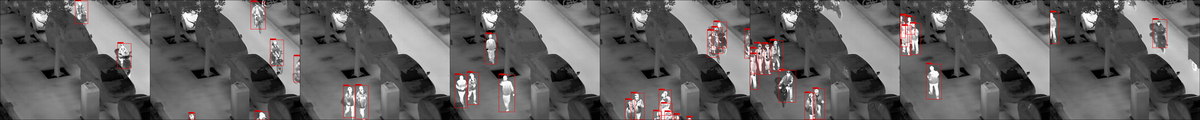} \\

    \makebox[0pt][r]{\makebox[10pt]{\raisebox{15pt}{\rotatebox[origin=c]{90}{\scriptsize $\text{ModTr}_{\odot}$}}}}
    \includegraphics[width=0.95\textwidth]{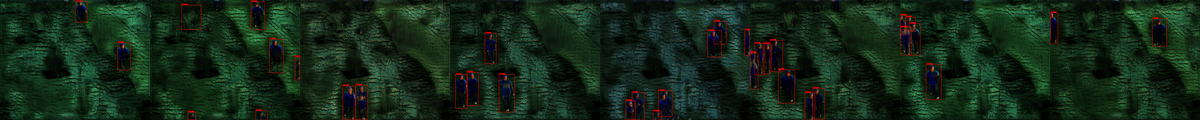} \\

    \midrule
    FLIR Test Dataset \\
    \midrule

    \makebox[0pt][r]{\makebox[10pt]{\raisebox{15pt}{\rotatebox[origin=c]{90}{\scriptsize GT}}}}
    \includegraphics[width=0.95\textwidth]{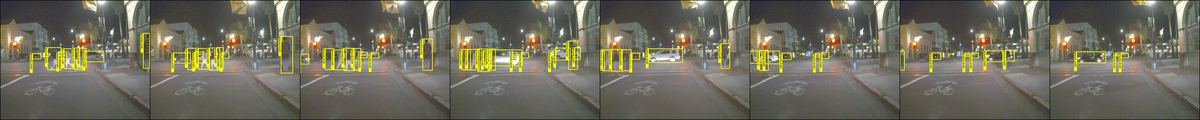} \\

    \makebox[0pt][r]{\makebox[10pt]{\raisebox{15pt}{\rotatebox[origin=c]{90}{\scriptsize FT}}}}
    \includegraphics[width=0.95\textwidth]{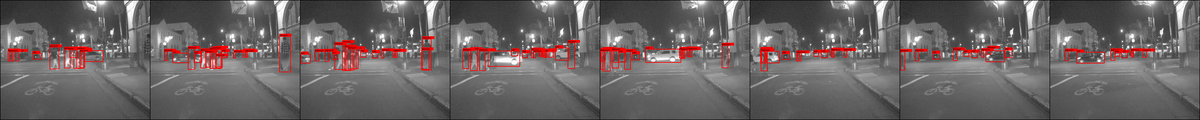} \\

    \makebox[0pt][r]{\makebox[10pt]{\raisebox{15pt}{\rotatebox[origin=c]{90}{\scriptsize $\text{ModTr}_{\odot}$}}}}
    \includegraphics[width=0.95\textwidth]{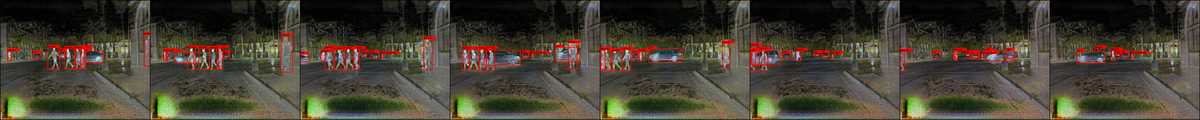}  \\

    \bottomrule

    \end{tabular}
\caption{Illustration of a sequence of $8$ images of LLVIP and FLIR dataset for Faster R-CNN. For each dataset, the first row is the RGB modality, followed by the IR modality and different representations created by ModTr. For visualizations of other detectors and variants of ModTr, please refer to the supplementary materials.}

\label{fig:04.04_qualitative_experiments}

\end{figure*}

\subsection{Knowledge Preservation through Input Modality Translation}

ModTr is designed to prevent catastrophic forgetting by keeping the weights of the pre-trained detector fixed. In this section, we demonstrate how various adaptation paradigms, shown in \Cref{fig:different_methods}, effectively solve the final task while preserving intrinsic knowledge. We compare our proposed method, ModTr, with two fine-tuning baseline methods. The first baseline method involves N-detectors, each fine-tuning the target modality individually. The second baseline method employs a single detector trained on the joint modality using balanced sampling. Note that while a copy of the original detector can be used in the N-detectors paradigm, it is unavailable in the $1$-detector paradigm because the original modality is assumed to be inaccessible during training.

In all scenarios, we use COCO as the pre-training dataset and LLVIP and FLIR as target domains. Specifically, in the N-detectors scenario (Fig.\ref{fig:different_methods}a), we fine-tune one detector on each dataset and use a copy of the original detector for the RGB modality. In the $1$-detector scenario (Fig.\ref{fig:different_methods}c), we fine-tune one detector on the combined FLIR and LLVIP datasets. In the N-ModTr-1-Detector scenario (Fig.\ref{fig:different_methods}b), two translators are trained, one per dataset. To assess catastrophic forgetting, we re-evaluate each scenario on COCO-val.

\Cref{tab:modality_translation} shows the final performance. While all adaptation paradigms achieve relatively similar performance, the $1$-detector method completely fails in the zero-shot scenario. The N-detectors method mitigates this by duplicating the detector three times. In contrast, ModTr preserves knowledge using a single detector and three efficient translators, demonstrating its practicality for embedded devices, as it requires less memory. Based on the average performance on all datasets, ModTr obtains the best results.

\begin{table}[h]
\caption{Detection performance (AP) of knowledge preserving techniques N-Detectors, 1-Detector, and N-ModTr-1-Detector, using three different detectors (FCOS, RetinaNet, and Faster R-CNN). The methods were evaluated on COCO and IR test sets of LLVIP and FLIR datasets.}
\label{tab:modality_translation}
    \centering
    \resizebox{0.9\textwidth}{!}{%
    \begin{tabular}{lqqqq}
        \toprule

        \rowcolor{white}
        {\textbf{Detector}} & \textbf{Dataset} & {\textbf{N-Detectors}} &  {\textbf{1-Detector}} &  {\textbf{N-ModTr-1-Det.}} \\
        \midrule

        \rowcolor{white}
        \multirow{2}{*}[-0.8em]{\textbf{FCOS}} & LLVIP & 57.37 ± 2.19 & \textbf{58.55 ± 0.89} & 57.63 ± 0.66 \\
        
         & FLIR & 27.97 ± 0.59 & 26.70 ± 0.48 &  \textbf{35.49 ± 0.94} \\
        \rowcolor{white}
        & COCO & \textbf{38.41 ± 0.00} & 00.33 ± 0.04 & \textbf{38.41 ± 0.00} \\
        \cmidrule{2-5}
        & \textbf{AVG.} & 41.25 ± 0.92 & 28.52 ± 0.47 & \textbf{43.84 ± 0.53} \\
        
        \midrule
        \rowcolor{white}
        \multirow{2}{*}[-0.8em]{\textbf{RetinaNet}} & LLVIP & 53.79 ± 1.79 & 53.26 ± 3.02 & \textbf{54.83 ± 0.61} \\
        
         & FLIR & 28.46 ± 0.50 & 25.19 ± 0.72 & \textbf{34.27 ± 0.27} \\
        \rowcolor{white}
        & COCO & \textbf{35.48 ± 0.00} & 00.29 ± 0.01 & \textbf{35.48 ± 0.00} \\
        \cmidrule{2-5}
        & \textbf{AVG.} & 39.24 ± 0.76 & 26.24 ± 1.28 & \textbf{41.52 ± 0.29} \\
        
        \midrule
        \rowcolor{white}
        \multirow{2}{*}[-0.8em]{\textbf{Faster R-CNN}} & LLVIP & 59.62 ± 1.23 & \textbf{62.50 ± 1.29} & 57.97 ± 0.85 \\
         & FLIR & 30.93 ± 0.46 & 28.90 ± 0.33 &  \textbf{37.21 ± 0.46} \\
        \rowcolor{white}
        & COCO &  \textbf{39.78 ± 0.00} & 00.40 ± 0.00 & \textbf{39.78 ± 0.00} \\
        \cmidrule{2-5}
        & \textbf{AVG.} & 43.44 ± 0.56 & 30.60 ± 0.54 & \textbf{44.98 ± 0.43} \\
     
        \bottomrule
    \end{tabular}
    }
\end{table}

\subsection{Visualization of ModTr Translated Images}

In~\fref{fig:04.04_qualitative_experiments}, we present qualitative results for LLVIP and FLIR, alongside a comparison with fine-tuning. Each dataset section includes three rows: the first row displays the ground-truth RGB images, the second row showcases the results of fine-tuning using IR, and the last row features ModTr with a Hadamard product-based fusion over the Faster R-CNN detector. Due to space constraints, additional visualizations for other detectors and fusion strategies are provided in the supplementary materials. Notably, the IR results exhibit some false positives, particularly when detected objects overlap. Our method mitigates this issue effectively. Further insights, provided in the supplementary materials, reveal how our method effectively blurs or removes objects that do not belong to the target classes, thereby enhancing detection accuracy. Although the obtained intermediate representations are not visually pleasant, they prove more efficient for incorporating the knowledge necessary for the OD. Additionally, we conducted experiments with loss function terms aimed at enhancing the visual effects of the image, but they were not conclusive in terms of helping the detection performance.

\subsection{Fine-tuning of ModTr and the Detector}

\begin{figure}[!htb]
  \centering
  \includegraphics[width=0.8\linewidth]{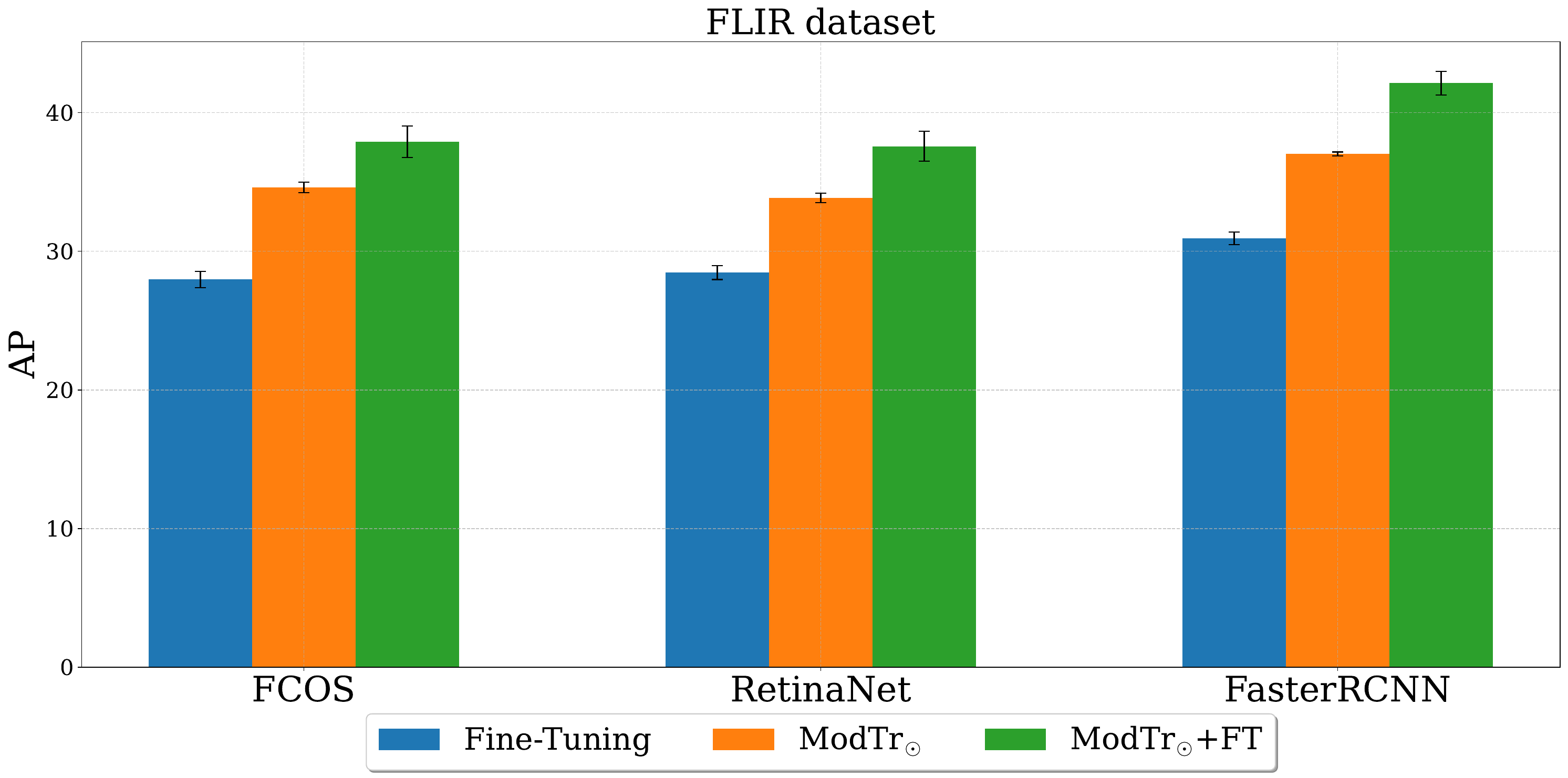}
  \caption{Comparison of the performance of fine-tuning the ModTr and normal fine-tuning on the FLIR dataset for the three different detectors (FCOS, RetinaNet, and Faster R-CNN). In blue, the Fine-tuning; in orange, the ModTr$_{\odot}$, and in green, ModTr$_{\odot}+\text{FT}$.}
  \label{fig:ft}
\end{figure}

The main reason to use ModTr is to avoid fine-tuning the detector for a specific task so that it can preserve its knowledge and be used for multiple modalities. However, in this section, we consider what would happen if we learn jointly ModTr and the detector weights.
Results are reported in~\fref{fig:ft}. We see that fine-tuning the detector can further boost performance. Thus, another application of ModTr could be used to improve the fine-tuning of a detector with a reduced additional computational cost.

\section{Conclusion} \label{sec:conclusion}

In this paper, a novel method called ModTr is proposed for adapting RGB object detectors (ODs) for IR modality without changing their parameters. A key advantage of our approach is that it preserves the full knowledge of the detector, allowing the translation network to act as a node that changes the modality for an unaltered detector. This is much more flexible and computationally efficient than having a specialized OD for each modality. Our approach performs well in various settings, outperforming powerful image-to-image models and previous competitors. We evaluated ModTr for different tasks, including detection based on image translation, comparison with traditional fine-tuning, and incremental IR modality application. Experimental results show the high performance and versatility of our method in all these settings.

Additionally, to explore integrating modalities beyond IR, we applied ModTr to Canny edges extracted from IR images as detailed in the supplementary material. While ModTr significantly enhances the performance of zero-shot RGB OD on edges, it still does not match the effectiveness of full fine-tuning on this other modality. We believe this shortfall arises from the limited information provided by edges compared to the richer data in the IR modality, leading to lower initial zero-shot OD performance. A potential solution is to replace the deterministic translator module within ModTr with a generative model. This substitution could enrich modality information by generating the missing data, potentially improving the zero-shot detector's performance. This promising direction will be explored in future research.

\section*{Acknowledgments} This work was supported in part by Distech Controls Inc., the Natural Sciences and Engineering Research Council of Canada, the Digital Research Alliance of Canada, and MITACS.

\title{Supplementary Material: {\color{red}Mod}ality {\color{red}Tr}anslation for Object Detection Adaptation Without Forgetting Prior Knowledge}

\titlerunning{ModTr for Object Detection}

\author{Heitor Rapela Medeiros\thanks{Email: heitor.rapela-medeiros.1@ens.etsmtl.ca}, Masih Aminbeidokhti, \\ Fidel Guerrero Pena, David Latortue, \\ Eric Granger, and Marco Pedersoli}

\institute{LIVIA, Dept. of Systems Engineering. ETS Montreal, Canada}

\authorrunning{Medeiros et al.}

\maketitle

In this supplementary material, we provide additional information to reproduce our work. The code is available at \url{https://github.com/heitorrapela/ModTr}. This supplementary material is divided into the following sections: Detailed diagrams (Section~\ref{sec:detailed_diagrams}), Different fusion strategies (Section~\ref{sec:different_fusion_strategies}), Quantitative Results (Section~\ref{sec:quantitative_results}), in which we provide additional numerical results in terms of APs, Qualitative Results (Section~\ref{sec:qualitative_results}), in which we provide additional visualizations and Additional Modality: Canny Edges (Section~\ref{sec:canny_edge}), in which we show experiments with edges extracted from IR images using Canny edge detector.

\section{Detailed Diagrams}
\label{sec:detailed_diagrams}

\begin{figure}[!htp]
    \centering
    \includegraphics[width=0.50\columnwidth]{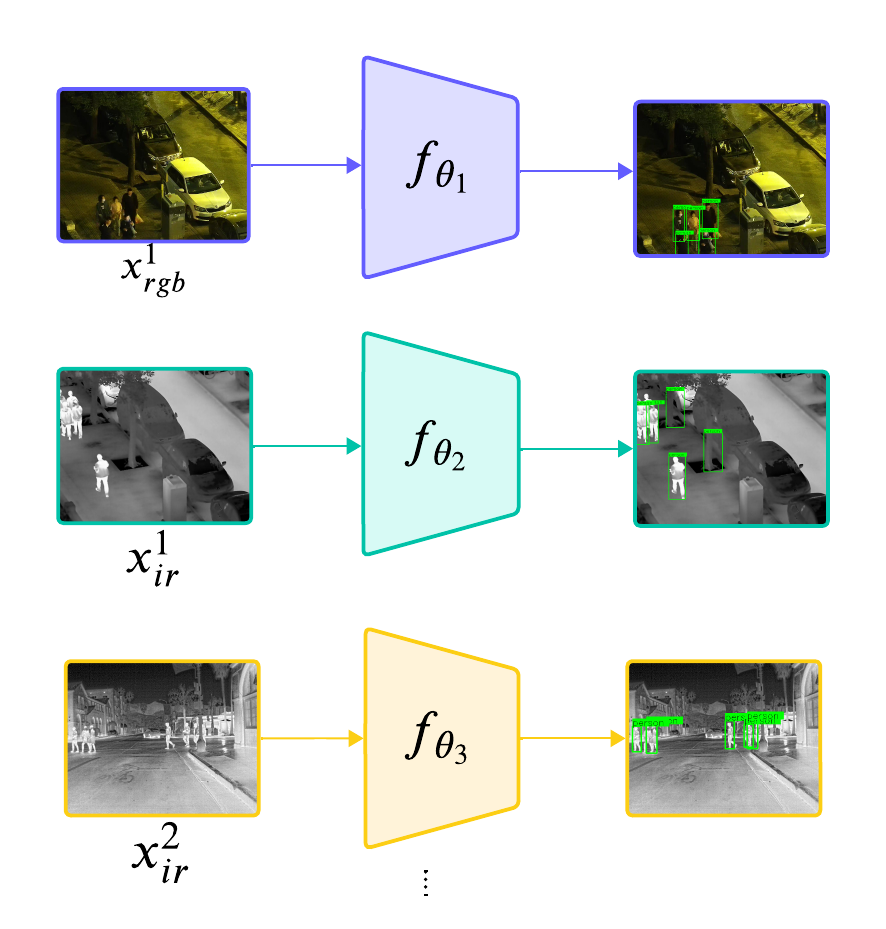}
    \caption{The simplest approach is to use a different detector adapted to each modality. This can lead to a high level of accuracy but requires storing models in memory multiple times. In purple is the RGB detector, in green is one IR detector for one dataset, and in yellow is another detector for another IR dataset.}
    \label{fig:N-Detectors}
\end{figure}

In this section, we expand and detail the diagrams provided in the main manuscript. In~\fref{fig:N-Detectors}, we describe the traditional approach of employing specialized detectors for individual modalities. For example, we depict an RGB detector (highlighted in purple) and two IR detectors (highlighted in green and yellow), with each trained over a different dataset.

\begin{figure}[!htp]
    \centering
    
    \includegraphics[width=0.8\columnwidth]{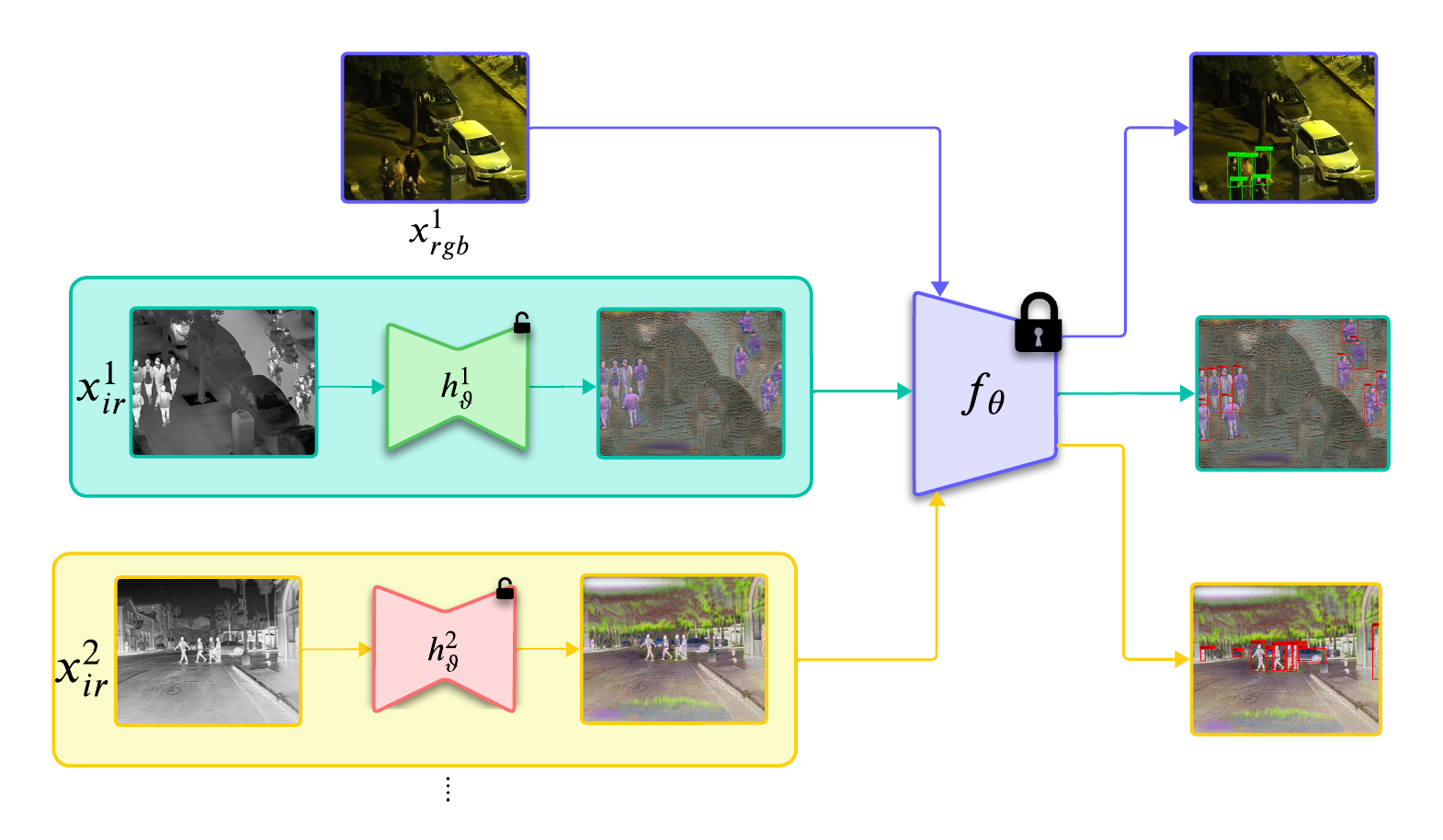}

    \caption{Our proposed solution is based on using a single pre-trained detector model normally trained on the more abundant data (RGB) and then adapting the input through our ModTr block.}
    \label{fig:N-ModTr-1-Detector}
\end{figure}

In~\fref{fig:N-ModTr-1-Detector}, we illustrate the proposed ModTr. This method involves using a single pre-trained detector model typically trained on the more prevalent data, i.e., RGB, and additional input adaptation network. For clarity, the RGB modality (along with the RGB detector) is depicted in purple, while an adaptation block of ModTr is shown in green for one IR modality and in red for the other IR modality with different distribution.

\begin{figure}[!htp]
    \centering
    
    \includegraphics[width=0.5\columnwidth]{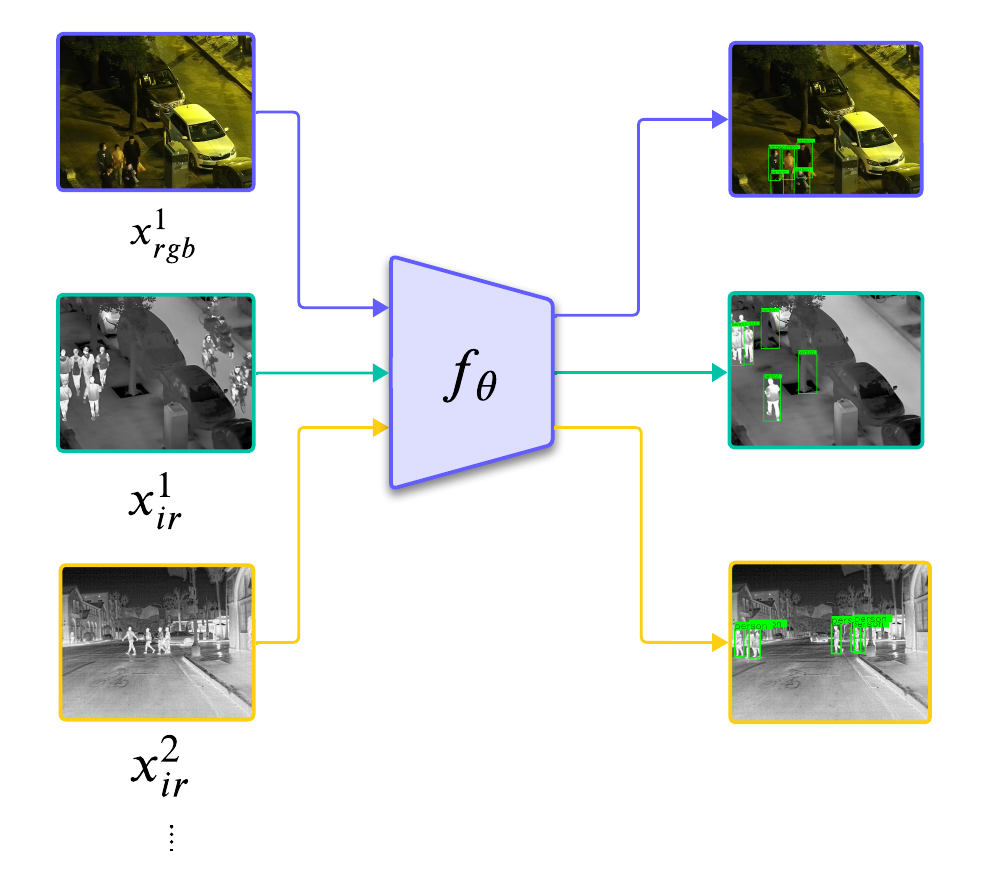}

    \caption{A single detector is trained on all modalities jointly. This allows the use of a single model but requires access to all modalities jointly, which is often not possible, especially when dealing with large pre-trained models.}
    \label{fig:1-Detector}
\end{figure}

Lastly, we present a final diagram (\fref{fig:1-Detector}), depicting a detector trained on the joint distribution of all modalities. The detector, shown in purple, undergoes training with all available modalities. While this approach enables the model to learn shared features, it may not be optimal. Nevertheless, it incurs a lower memory cost compared to employing one detector for each modality.

\section{Different fusion strategies}
\label{sec:different_fusion_strategies}

Here, we present a comprehensive overview of the alternative fusion strategies we explored. In addition to the element-wise product fusion, which produced the best results for our study and was reported in the main manuscript, we also tested strategies \textbf{ModTr$_{+}$}, and \textbf{ModTr$_{\oplus}$}, detailed below: \\

\textbf{ModTr$_{+}$}: The addition mechanism involves forwarding the input modality and summing it with the output of the translation network, as detailed in~\eref{eq:modtr_addition}. This residual connection of the input serves as regularization for the model, aiding the network in learning the missing information necessary for detector operation. This operator learns the new representation by amplifying pixel values when the weights of the translation representation tend toward $1$, or preserving the original information when they tend toward $0$. Such a range of values is due to our modification on the U-Net~\cite{ronneberger2015u} to generate, in which we changed the last layer to a Sigmoid function layer so we can better control the generated image to be closer to a real RGB-like image:

\begin{equation} 
\begin{split}
\mathcal{L}_{\text{ModTr}_{+}}(x, Y; \vartheta) & = \mathcal{L}_{det}(f_\theta\left(h_\vartheta^d(x)+x\right), Y).
\end{split}
\label{eq:modtr_addition}
\end{equation}

\textbf{ModTr$_{\oplus}$}: The subsequent fusion mechanism draws inspiration from DenseFuse \cite{li2018densefuse}, which employs the relative importance of pixels as an attention mechanism for both the translation network's output and input. This attention mechanism operates by providing a weighted average of the channels. The implementation details for such an operator can be found on \cite{li2018densefuse}. Then, the proposed loss function is given by the following~\eref{eq:modtr_oplus}:

\begin{equation} 
\begin{split}
\mathcal{L}_{\text{ModTr}_{\oplus}}(x, Y; \vartheta) & = \mathcal{L}_{det}(f_\theta\left(\Phi(h_\vartheta^d(x), x)\right), Y),
\end{split}
\label{eq:modtr_oplus}
\end{equation}
with 
\begin{equation*}
\Phi(x,\hat{x})= \frac{x\odot e^x + \hat{x}\odot e^{\hat{x}}}{e^x+e^{\hat{x}}}.
\end{equation*}

\section{Quantitative Results}
\label{sec:quantitative_results}

In this section, we provide further details for the different AP metrics of the experiments in the main manuscript. In~\tref{tab:img2img}, we compare the best ModTr with different image translation strategies. Some of the competitors use RGB data for the translation, while others utilize bounding box annotations. Here, we can see that ModTr outperforms the competitors in terms of AP metrics across different detectors, demonstrating its superiority in terms of classification and localization. For instance, when compared with HalluciDet~\cite{medeiros2024hallucidet} using FCOS, which was the second best, our method shows an improvement of approximately $28$ better in terms of AP$_{50}$, with a similar trend observed for RetinaNet on the LLVIP dataset. The gap is narrower with Faster R-CNN, with an improvement of only around $2$ AP$_{50}$. While the gap is smaller for the FLIR dataset, it remains consistent across all APs and detectors; for example, an improvement of $6$ AP$_{50}$ for Faster R-CNN, approximately $14$ AP$_{50}$ for RetinaNet, and $11$ AP$_{50}$ for FCOS. In terms of localization for the FLIR dataset, significant improvements are observed, such as around $7$ AP for Faster R-CNN, $11$ AP for RetinaNet, and $11$ AP for FCOS when compared with HalluciDet. For the other methods, which do not rely on bounding boxes, substantial improvements are evident; for instance, our method exhibits an improvement of more than $11$ AP over FastCUT~\cite{park2020contrastive}, with similar trends observed for other competitors.

\begin{table}[!htp]
\caption{Detection performance of ModTr versus baseline image-to-image methods to translate the IR to RGB-like images, using three different detectors (FCOS, RetinaNet, and Faster R-CNN).
The methods were evaluated on IR test set of LLVIP and FLIR datasets. The RGB column indicates if the method required access to RGB images during training, and Box refers to the use of ground truth bboxes during training.
}
        \centering
        \resizebox{1.0\columnwidth}{!}{%
        \begin{tabular}{lccccccccccccccc}
    
            \toprule
            
            \multirow{3}{*}[-1em]{\textbf{Image translation}} & \multirow{3}{*}[-1em]{\textbf{RGB}} & \multirow{3}{*}[-1em]{\textbf{Box}}  & \multicolumn{8}{c}{\textbf{Test Set IR (Dataset: LLVIP)}} \\
            \cmidrule(lr){4-13}
            \addlinespace[5pt]

            {} & {} & {} & \multicolumn{3}{c}{\multirow{2}{*}[1em]{\textbf{FCOS}}} & \multicolumn{3}{c}{\multirow{2}{*}[1em]{\textbf{RetinaNet}}} & \multicolumn{3}{c}{\multirow{2}{*}[1em]{\textbf{Faster R-CNN}}} \\
    
            \cmidrule(lr){4-6}
            \cmidrule(lr){7-9}
            \cmidrule(lr){10-13}
            \addlinespace[5pt]
            
            {} &  {} &  {} &  \textbf{AP$_{50}\uparrow$} & \textbf{AP$_{75}\uparrow$} & \textbf{AP${}\uparrow$} &   \textbf{AP$_{50}\uparrow$} & \textbf{AP$_{75}\uparrow$} & \textbf{AP${}\uparrow$} & \textbf{AP$_{50}\uparrow$} & \textbf{AP$_{75}\uparrow$} & \textbf{AP${}\uparrow$} & \\

            \midrule
            
            Histogram Equal.~\cite{herrmann2018cnn}& {} & {} & 53.74 ± 0.00 & 32.57 ± 0.00 & 31.69 ± 0.00 & 59.93 ± 0.00 & 33.04 ± 0.00 & 33.16 ± 0.00 & 65.70 ± 0.04 & 39.02 ± 0.11 & 38.33 ± 0.02 \\

            \rowcolor[HTML]{EFEFEF}
            CycleGAN~\cite{zhu2017unpaired}& \checkmark & {} & 41.72 ± 1.63 & 23.83 ± 0.78 & 23.85 ± 0.76 & 43.17 ± 1.52 & 22.34 ± 0.88 & 23.34 ± 0.53 & 45.44 ± 1.89 & 26.82 ± 1.59 & 26.54 ± 1.20 \\
    
            CUT~\cite{park2020cut}& \checkmark & {} &  26.48 ± 2.88 & 13.68 ± 2.72 & 14.30 ± 2.25 & 25.64 ± 3.77 &  11.74 ± 2.33 &  13.12 ± 2.07 & 27.96 ± 1.70 & 13.59 ± 2.77 & 14.78 ± 1.82 \\
    
            \rowcolor[HTML]{EFEFEF}
            FastCUT~\cite{park2020cut}& \checkmark & {} & 34.92 ± 3.63  & 19.07 ± 1.33 & 19.39 ± 1.52 & 35.73 ± 2.53 & 16.36 ± 0.44 & 18.11 ± 0.79 & 42.09 ± 3.51 & 21.44 ± 1.57 & 22.91 ± 1.68 \\
            
            HalluciDet~\cite{medeiros2024hallucidet}& \checkmark &  \checkmark &  64.17 ± 0.61 & 18.80 ± 1.45 & 28.00 ± 0.92 & 60.38 ± 3.59 & 06.75 ± 1.38 & 19.95 ± 2.01 & 90.07 ± 0.72 & 51.23 ± 1.81 & 57.78 ± 0.97 \\
            
            \midrule
     
            \rowcolor[HTML]{EFEFEF}
             ModTr$_{\odot}$ (ours)  & {} &  \checkmark & \textbf{92.04 ± 0.47} & \textbf{63.84 ± 0.93} & \textbf{57.63 ± 0.66} & \textbf{91.56 ± 0.64} & \textbf{59.49 ± 1.11} & \textbf{54.83 ± 0.61} & \textbf{91.82 ± 0.49} & \textbf{62.51 ± 0.87} & \textbf{57.97 ± 0.85} \\
    
            \toprule
            
            \multirow{3}{*}[-1em]{\textbf{Image translation}} & \multirow{3}{*}[-1em]{\textbf{RGB}} & \multirow{3}{*}[-1em]{\textbf{Box}}  & \multicolumn{8}{c}{\textbf{Test Set IR (Dataset: FLIR)}} \\
            \cmidrule(lr){4-13}
            \addlinespace[5pt]

            {} & {} & {} & \multicolumn{3}{c}{\multirow{2}{*}[1em]{\textbf{FCOS}}} & \multicolumn{3}{c}{\multirow{2}{*}[1em]{\textbf{RetinaNet}}} & \multicolumn{3}{c}{\multirow{2}{*}[1em]{\textbf{Faster R-CNN}}} \\
    
            \cmidrule(lr){4-6}
            \cmidrule(lr){7-9}
            \cmidrule(lr){10-13}
            \addlinespace[5pt]
            
            {} &  {} &  {} &  \textbf{AP$_{50}\uparrow$} & \textbf{AP$_{75}\uparrow$} & \textbf{AP${}\uparrow$} &   \textbf{AP$_{50}\uparrow$} & \textbf{AP$_{75}\uparrow$} & \textbf{AP${}\uparrow$} & \textbf{AP$_{50}\uparrow$} & \textbf{AP$_{75}\uparrow$} & \textbf{AP${}\uparrow$} & \\

            \midrule

            Histogram Equal.~\cite{herrmann2018cnn} & {} & {} &  52.09 ± 0.00 & 16.44 ± 0.00 & 22.76 ± 0.00 & 53.13 ± 0.00 & 16.50 ± 0.00 & 23.06 ± 0.00 & 56.50 ± 0.10 & 17.62 ± 0.04 & 24.61 ± 0.01 \\

            \rowcolor[HTML]{EFEFEF}
            CycleGAN~\cite{zhu2017unpaired}& \checkmark & {} &  49.01 ± 1.28 & 21.16 ± 0.71 & 23.92 ± 0.97 & 49.04 ± 1.71 & 19.93 ± 0.54 & 23.71 ± 0.70 & 54.48 ± 2.09 & 23.08 ± 1.54 & 26.85 ± 1.23 \\
    
            CUT~\cite{park2020cut}& \checkmark & {} & 38.70 ± 1.05 & 14.85 ± 0.49 & 18.16 ± 0.75 & 39.08 ± 1.42 & 13.69 ± 0.61 & 17.84 ± 0.75 & 43.34 ± 1.53 &  16.09 ± 0.38 & 20.29 ± 0.48 \\
    
            \rowcolor[HTML]{EFEFEF}
            FastCUT~\cite{park2020cut}& \checkmark & {} &  45.19 ± 4.46 & 22.93 ± 2.09 & 24.02 ± 2.37 & 43.04 ± 4.95 & 19.82 ± 2.78 & 22.00 ± 2.73 & 49.98 ± 4.57 & 25.52 ± 2.85 & 26.68 ± 2.59 \\
    
            HalluciDet~\cite{medeiros2024hallucidet} & \checkmark & \checkmark &  54.20 ± 2.50  &  17.36 ± 2.23 &  23.74 ± 2.09 & 52.06 ± 1.47 & 16.21 ± 0.31 & 22.29 ± 0.45 & 63.11 ± 1.54 & 23.91 ± 1.10 & 29.91 ± 1.18 \\
            
            \midrule
    
            \rowcolor[HTML]{EFEFEF}
             ModTr$_{\odot}$ (ours)  & {} & \checkmark &  \textbf{65.99 ± 0.78} & \textbf{33.73 ± 1.74} & \textbf{35.49 ± 0.94} & \textbf{66.31 ± 0.93} & \textbf{31.22 ± 0.69} & \textbf{34.27 ± 0.27} & \textbf{69.20 ± 0.36} & \textbf{34.58 ± 0.56} & \textbf{37.21 ± 0.46} \\

            \bottomrule
        \end{tabular}
        }
        \label{tab:img2img}
    
    \end{table}

In~\tref{tab:modtr_ft}, we show that compared with fine-tuning (FT), the ModTr is better even without modifying the parameters of the detector. Thus, it preserves the detector's knowledge for further tasks while improving performance. For instance, in terms of localization with AP$_{75}$ and AP for FCOS, RetinaNet on LLVIP, we reached the performance of the FT, while with AP$_{50}$, we were comparable. For the FLIR dataset, we outperform the FT in all detectors for the different APs and also in all different fusion strategies.

\begin{table}[!htp]
\caption{Detection performance (AP) of ModTr with different fusion strategies versus baseline fine-tuning (FT) of the detector, using three different detectors (FCOS, RetinaNet, and Faster R-CNN. The methods were evaluated on IR test set of LLVIP and FLIR datasets.}
    \centering
    \resizebox{1.0\columnwidth}{!}{%
    \begin{tabular}{lccccccccccccc}

        \toprule

        \multirow{3}{*}[-1em]{\textbf{Method}} & \multicolumn{9}{c}{\textbf{Test Set IR (Dataset: LLVIP)}} \\
        \cmidrule(lr){2-11}
        \addlinespace[5pt]

        {}  & \multicolumn{3}{c}{\multirow{2}{*}[1em]{\textbf{FCOS}}} & \multicolumn{3}{c}{\multirow{2}{*}[1em]{\textbf{RetinaNet}}} & \multicolumn{3}{c}{\multirow{2}{*}[1em]{\textbf{Faster R-CNN}}} \\

        \cmidrule(lr){2-4}
        \cmidrule(lr){5-7}
        \cmidrule(lr){8-11}
        \addlinespace[5pt]
        
        {} &  \textbf{AP$_{50}\uparrow$} & \textbf{AP$_{75}\uparrow$} & \textbf{AP${}\uparrow$} &   \textbf{AP$_{50}\uparrow$} & \textbf{AP$_{75}\uparrow$} & \textbf{AP${}\uparrow$} & \textbf{AP$_{50}\uparrow$} & \textbf{AP$_{75}\uparrow$} & \textbf{AP${}\uparrow$} & \\

        \midrule

        Fine-Tuning & \textbf{93.14 ± 1.06} & 62.08 ± 3.37 & 57.37 ± 2.19 & \textbf{93.61 ± 0.59} & 56.20 ± 3.78 & 53.79 ± 1.79 & \textbf{94.64 ± 0.84} & 62.50 ± 0.79 &  \textbf{59.62 ± 1.23} \\

        \midrule
 
        \rowcolor[HTML]{EFEFEF}
        ModTr$_{+}$ & 90.57 ± 1.46 & 62.38 ± 0.31 & 56.44 ± 0.75 & 91.09 ± 0.73 & 55.06 ± 1.81 & 53.18 ± 1.03 & 91.89 ± 0.39 & 61.44 ± 0.72 & 57.14 ± 0.50 \\

        ModTr$_{\oplus}$ & 91.11 ± 0.84 & 62.69 ± 1.53 & 57.01 ± 0.71 & 90.49 ± 1.11 & 58.73 ± 0.55 & 54.43 ± 0.35 & 91.20 ± 0.46 & 61.31 ± 0.73 & 56.95 ± 0.37 \\
        
        \rowcolor[HTML]{EFEFEF}
        ModTr$_{\odot}$ & 92.04 ± 0.47 & \textbf{63.84 ± 0.93} & \textbf{57.63 ± 0.66} & 91.56 ± 0.64 & \textbf{59.49 ± 1.11} & \textbf{54.83 ± 0.61} & 91.82 ± 0.49 & \textbf{62.51 ± 0.87} & 57.97 ± 0.85 \\
    
        \bottomrule
        
        \multirow{3}{*}[-1em]{\textbf{Method}} & \multicolumn{9}{c}{\textbf{Test Set IR (Dataset: FLIR)}} \\
        \cmidrule(lr){2-11}
        \addlinespace[5pt]

        {}  & \multicolumn{3}{c}{\multirow{2}{*}[1em]{\textbf{FCOS}}} & \multicolumn{3}{c}{\multirow{2}{*}[1em]{\textbf{RetinaNet}}} & \multicolumn{3}{c}{\multirow{2}{*}[1em]{\textbf{Faster R-CNN}}} \\

        \cmidrule(lr){2-4}
        \cmidrule(lr){5-7}
        \cmidrule(lr){8-11}
        \addlinespace[5pt]
        
        {} &  \textbf{AP$_{50}\uparrow$} & \textbf{AP$_{75}\uparrow$} & \textbf{AP${}\uparrow$} &   \textbf{AP$_{50}\uparrow$} & \textbf{AP$_{75}\uparrow$} & \textbf{AP${}\uparrow$} & \textbf{AP$_{50}\uparrow$} & \textbf{AP$_{75}\uparrow$} & \textbf{AP${}\uparrow$} & \\

        \midrule

        Fine-Tuning & 60.22 ± 0.97 & 21.94 ± 0.42 & 27.97 ± 0.59 & 61.77 ± 1.02 & 22.37 ± 0.45 & 28.46 ± 0.50 & 66.15 ± 0.94 & 24.48 ± 0.71 & 30.93 ± 0.46 \\
        
        \midrule

        \rowcolor[HTML]{EFEFEF}
        ModTr$_{+}$ & 64.90 ± 0.48 & 32.78 ± 0.27 & 34.63 ± 0.24 & 65.30 ± 0.66 & 30.00 ± 0.81 & 33.70 ± 0.59 & 68.64 ± 0.77 & 34.96 ± 0.90 & 37.09 ± 0.74 \\

        ModTr$_{\oplus}$ & \textbf{65.46 ± 0.61} & \textbf{33.21 ± 0.55} & 34.94 ± 0.52 & 63.87 ± 0.51 & \textbf{30.93 ± 0.38} & 33.72 ± 0.22 & 68.64 ± 1.29 & \textbf{35.48 ± 0.33} & 37.16 ± 0.47 \\

        \rowcolor[HTML]{EFEFEF}
        ModTr$_{\odot}$ & 65.25 ± 0.33 & 32.24 ± 0.95 & \textbf{35.49 ± 0.94} & \textbf{64.96 ± 0.68} & \textbf{30.93 ± 0.50} & \textbf{34.27 ± 0.27} & \textbf{68.84 ± 0.40} & 34.77 ± 0.22 & \textbf{37.21 ± 0.46} \\

        \bottomrule
    \end{tabular}
    }
\label{tab:modtr_ft}
\end{table}

In~\tref{tab:diff_backbones_llvip}, we explore the potential of achieving comparable performance in detection tasks while significantly reducing the number of parameters by employing a smaller backbone for the translation network. The MobileNet$_{v2}$ with only $6.6$ million additional parameters, achieves performance on par with ResNet$_{34}$ which have $24.4$ million parameters. This reduction in parameters is even more pronounced when compared with a new detector on the desired modality. For example, a new FCOS detector would require more $33.2$ million parameters. This trend is similar over the various detectors and remains consistent for the FLIR dataset (\tref{tab:diff_backbones_flir}).

\begin{table}[!htp]
\caption{Detection performance of ModTr with different backbones for the translation networks with different numbers of parameters, using three different detectors (FCOS, RetinaNet, and Faster R-CNN). The methods were evaluated on IR test set of LLVIP dataset.}
\label{tab:diff_backbones_llvip}

    \centering    
    \resizebox{0.9\columnwidth}{!}{%
    \begin{tabular}{qgggg}

        \toprule
        \rowcolor{white}
        {} & \multicolumn{3}{c}{\textbf{Test Set IR (Dataset: LLVIP)}} & {} \\
        
        \midrule
        \rowcolor{white}
        \multicolumn{1}{c}{\textbf{Method}}  & \textbf{Parameters} &  \textbf{AP$_{50}\uparrow$} & \textbf{AP$_{75}\uparrow$} & \textbf{AP${}\uparrow$} \\
        \midrule
        \addlinespace[5pt]
        \rowcolor{white}
        \multicolumn{1}{c}{\textbf{FCOS}} & 33.2 M & {} & {} & {}  \\
        \midrule 
         MobileNet$_{v3s}$ & + 3.1 M & 83.34 ± 1.76 & 53.34 ± 0.75 & 50.05 ± 1.01 \\
        \rowcolor{white}
        MobileNet$_{v2}$ & + 6.6 M & 90.36 ± 0.54 & 60.17 ± 0.56 & 55.33 ± 0.62 \\
        ResNet$_{18}$ & + 14.3 M & 89.53 ± 1.03 & 58.54 ± 1.57 & 54.25 ± 0.90 \\
        \rowcolor{white}
        ResNet$_{34}$ & + 24.4 M & 90.90 ± 1.29 & 62.96 ± 2.44 & 56.93 ± 1.44 \\

        \midrule
        \addlinespace[5pt]
        \rowcolor{white}
        \multicolumn{1}{c}{\textbf{RetinaNet}}  & 34.0 M & {} & {} & {} \\
        \midrule 
        MobileNet$_{v3s}$ & + 3.1 M & 87.67 ± 0.18 & 49.99 ± 0.57 & 49.65 ± 0.07 \\
        \rowcolor{white}
        MobileNet$_{v2}$ & + 6.6 M & 90.21 ± 0.82 & 54.60 ± 2.62 & 52.42 ± 1.33 \\
        ResNet$_{18}$ & + 14.3 M & 89.53 ± 1.82 & 52.68 ± 2.06 & 51.40 ± 1.40 \\
        \rowcolor{white}
        ResNet$_{34}$ & + 24.4 M  & 90.35 ± 0.60 & 57.18 ± 0.29 & 53.60 ± 0.41 \\

        \midrule
        \addlinespace[5pt]
        \rowcolor{white}
        \multicolumn{1}{c}{\textbf{Faster R-CNN}} & 41.8 M & {} & {} & {} \\
        \midrule 
        MobileNet$_{v3s}$ & + 3.1 M & 89.14 ± 0.63 & 56.85 ± 0.69 & 54.51 ± 0.28 \\
        \rowcolor{white}
        MobileNet$_{v2}$ & + 6.6 M & 91.32 ± 0.73 & 60.34 ± 0.66 & 56.15 ± 0.51  \\     
        ResNet$_{18}$ & + 14.3 M & 90.81 ± 0.46 & 59.39 ± 2.06 & 55.53 ± 1.14 \\
        \rowcolor{white}
        ResNet$_{34}$ & + 24.4 M  & 91.04 ± 0.29 & 60.53 ± 2.16 & 56.35 ± 0.65 \\

        \bottomrule
    \end{tabular}
    }
\end{table}

\begin{table}[!htp]
\caption{Detection performance (AP) of ModTr with different backbones for the translation networks with different numbers of parameters, using three different detectors (FCOS, RetinaNet, and Faster R-CNN). The methods were evaluated on IR test set of FLIR dataset.}
\label{tab:diff_backbones_flir}

    \centering   
    \resizebox{0.9\columnwidth}{!}{%
    \begin{tabular}{qgggg}

        \toprule
        \rowcolor{white}
        {} & \multicolumn{3}{c}{\textbf{Test Set IR (Dataset: FLIR)}} & {} \\

        \midrule
        \rowcolor{white}
        \multicolumn{1}{c}{\textbf{Method}}  & \textbf{Parameters} &  \textbf{AP$_{50}\uparrow$} & \textbf{AP$_{75}\uparrow$} & \textbf{AP${}\uparrow$} \\
        \midrule
        \addlinespace[5pt]
        \rowcolor{white}
        \multicolumn{1}{c}{\textbf{FCOS}}  & 33.2 M & {} & {} & {}  \\
        \midrule 
        
        MobileNet$_{v3s}$ & + 3.1 M & 56.73 ± 0.34 & 27.57 ± 0.65 & 29.66 ± 0.14 \\
        \rowcolor{white}
        MobileNet$_{v2}$ & + 6.6 M & 64.49 ± 0.99 & 32.17 ± 0.38 & 32.17 ± 0.38 \\
        ResNet$_{18}$ & + 14.3 M & 64.39 ± 1.68  & 32.72 ± 1.50 & 34.44 ± 1.13 \\
        \rowcolor{white}
        ResNet$_{34}$ & + 24.4 M  & 65.99 ± 0.78 & 33.73 ± 1.74 & 35.49 ± 0.94 \\

        \midrule
        \addlinespace[5pt]
        \rowcolor{white}
        \multicolumn{1}{c}{\textbf{RetinaNet}} & 34.0 M & {} & {} & {} \\
        \midrule 
        MobileNet$_{v3s}$ & + 3.1 M & 47.30 ± 0.54 & 18.63 ± 0.10 & 22.67 ± 0.18 \\
        \rowcolor{white}
        MobileNet$_{v2}$ & + 6.6 M & 64.01 ± 1.51 & 29.70 ± 0.62 & 33.12 ± 0.68  \\
        ResNet$_{18}$ & + 14.3 M & 64.20 ± 0.58  & 30.84 ± 0.47 & 33.44 ± 0.47 \\
        \rowcolor{white}
        ResNet$_{34}$ & + 24.4 M  & 66.31 ± 0.93 & 31.22 ± 0.69 & 34.27 ± 0.27 \\
        
        \midrule
        \addlinespace[5pt]
        \rowcolor{white}
        \multicolumn{1}{c}{\textbf{Faster R-CNN}} & 41.8 M & {} & {} & {} \\
        \midrule 
        MobileNet$_{v3s}$ & + 3.1 M & 61.03 ± 1.26 & 29.87 ± 0.86 & 32.06 ± 0.75 \\
        \rowcolor{white}
        MobileNet$_{v2}$ & + 6.6 M & 68.64 ± 0.56 & 34.76 ± 1.27 & 36.77 ± 0.67 \\
        ResNet$_{18}$ & + 14.3 M & 68.49 ± 0.53 & 34.52 ± 0.23 & 36.68 ± 0.22 \\
        \rowcolor{white}
        ResNet$_{34}$ & + 24.4 M  & 69.20 ± 0.36 & 34.58 ± 0.56 & 37.21 ± 0.46 \\

        \bottomrule
    \end{tabular}
    }
\end{table}

\section{Qualitative Results}
\label{sec:qualitative_results}

\begin{figure}[!htp]
\centering

\begin{tabular}{c}

    \toprule
    \textbf{Detector: FCOS} \\
    \midrule
      
    \begin{subfigure}[t]{0.23\textwidth}
        \caption{RGB - GT}
        \makebox[0pt][r]{\makebox[15pt]{\raisebox{30pt}{\rotatebox[origin=c]{90}{LLVIP}}}}%
        \includegraphics[width=\textwidth]
        {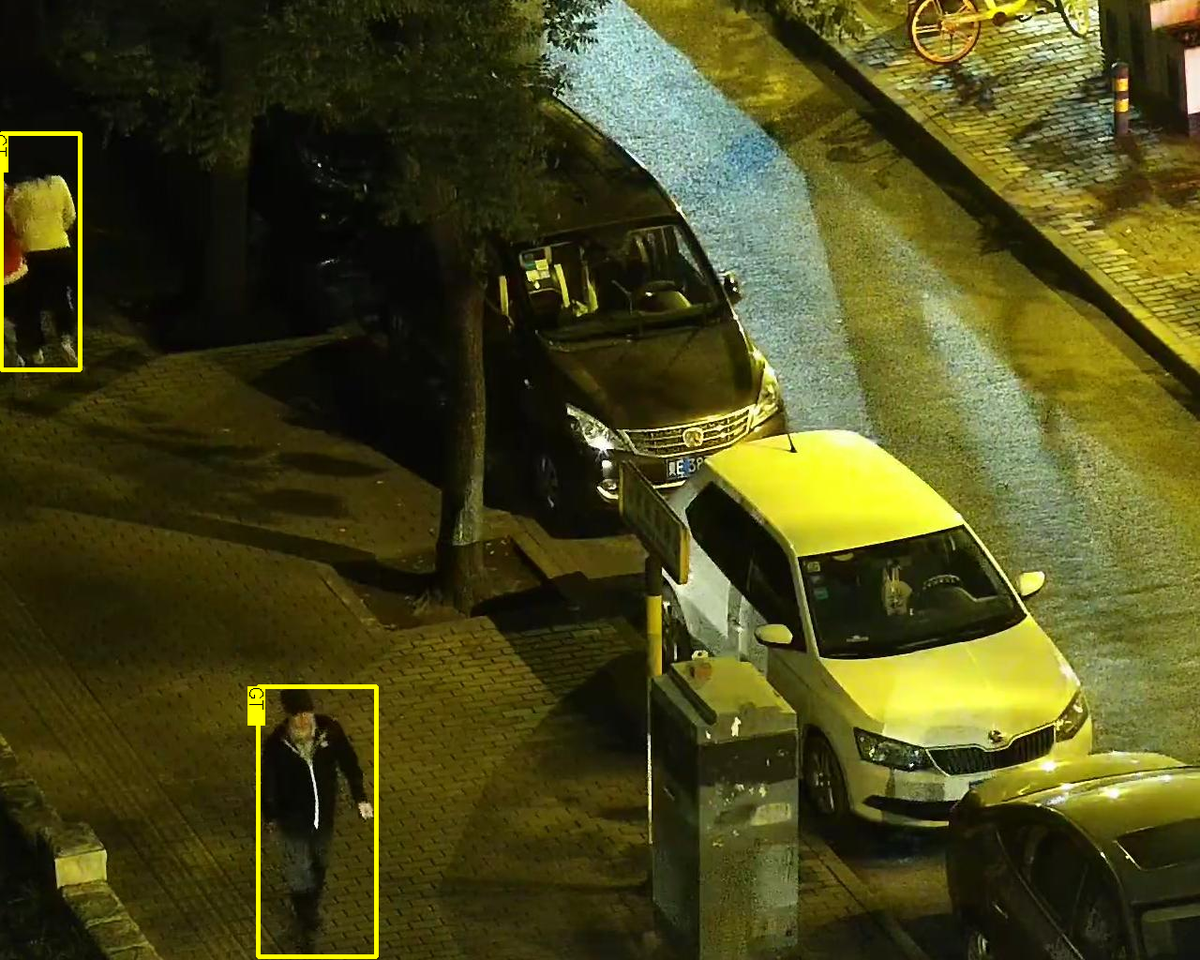}
        \makebox[0pt][r]{\makebox[15pt]{\raisebox{30pt}{\rotatebox[origin=c]{90}{FLIR}}}}%
        \includegraphics[width=\textwidth]
        {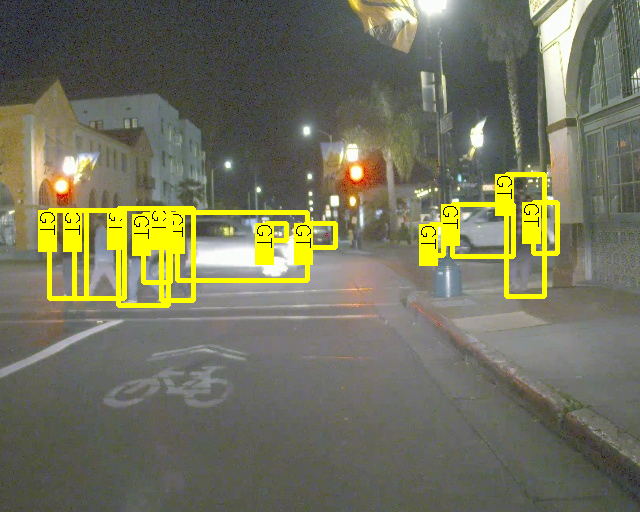}
    \end{subfigure}
    \begin{subfigure}[t]{0.23\textwidth}
        \caption{IR - FastCUT~\cite{park2020contrastive}}
        \includegraphics[width=\textwidth]  
        {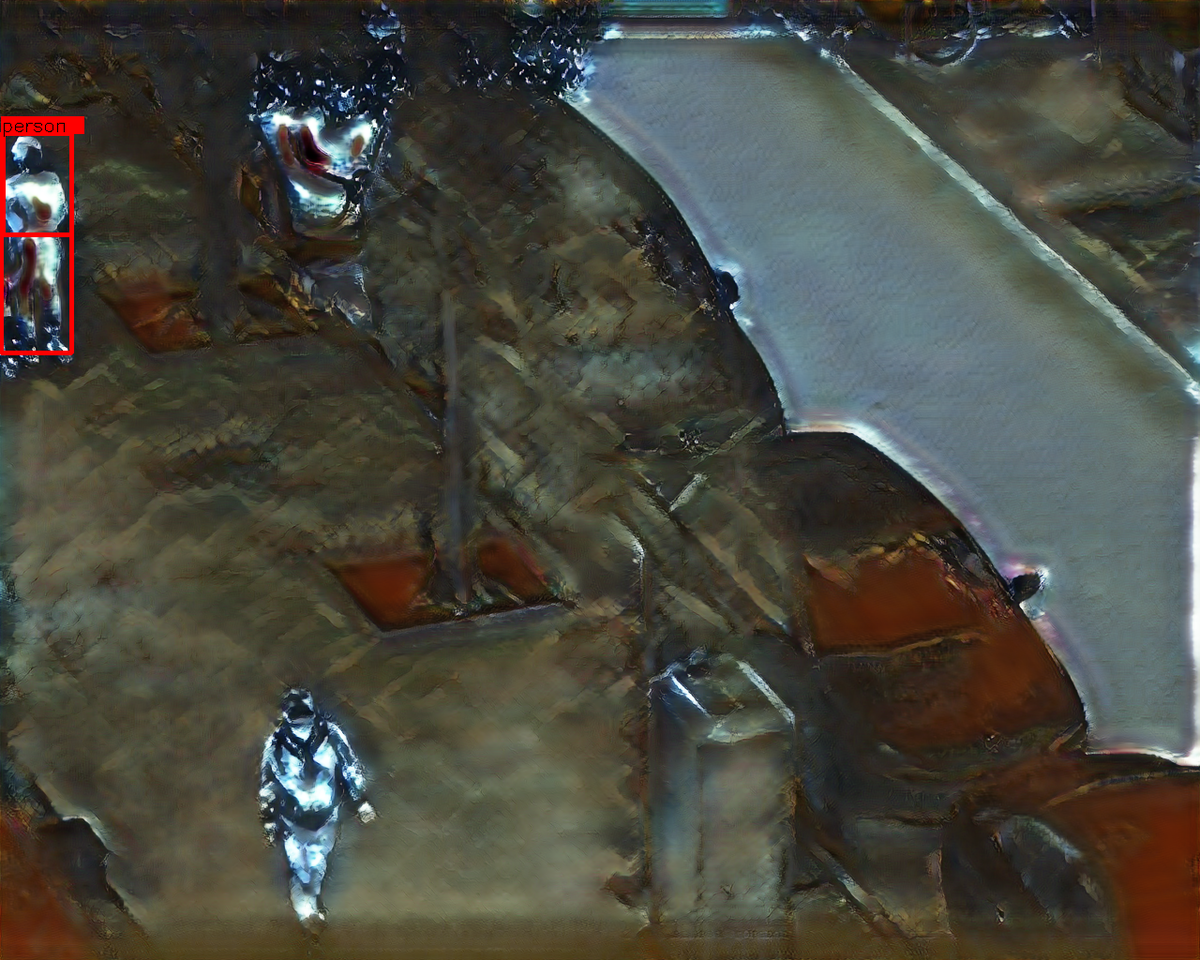}
        \includegraphics[width=\textwidth]
        {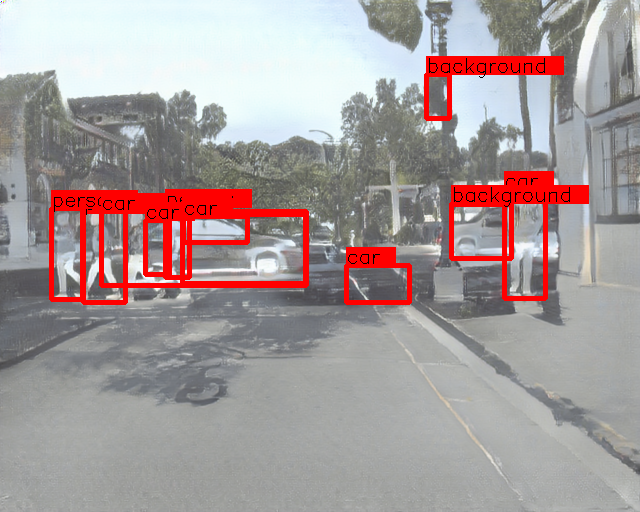}
    \end{subfigure}
    \begin{subfigure}[t]{0.23\textwidth}
        \caption{IR - Fine-tuning}
        \includegraphics[width=\textwidth]  
        {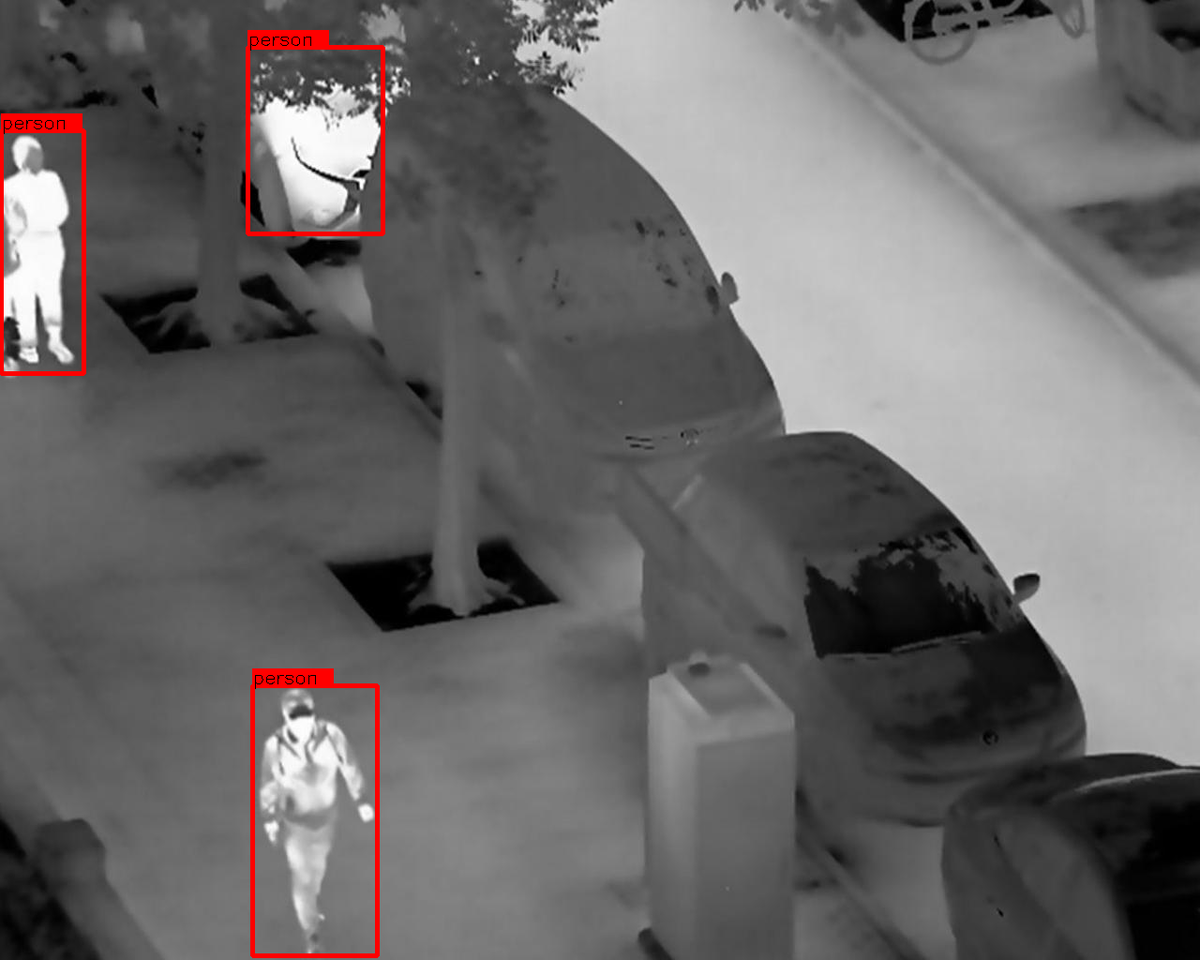}
        \includegraphics[width=\textwidth]
        {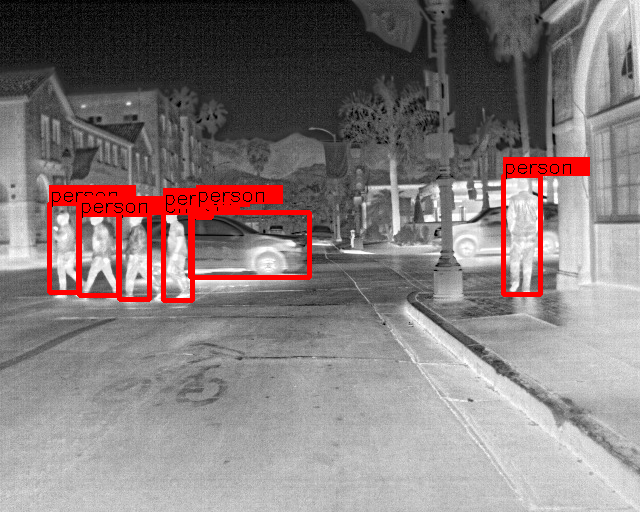}
    \end{subfigure}
    \begin{subfigure}[t]{0.23\textwidth}
        \caption{IR - ModTr (Ours)}
        \includegraphics[width=\textwidth]  
         {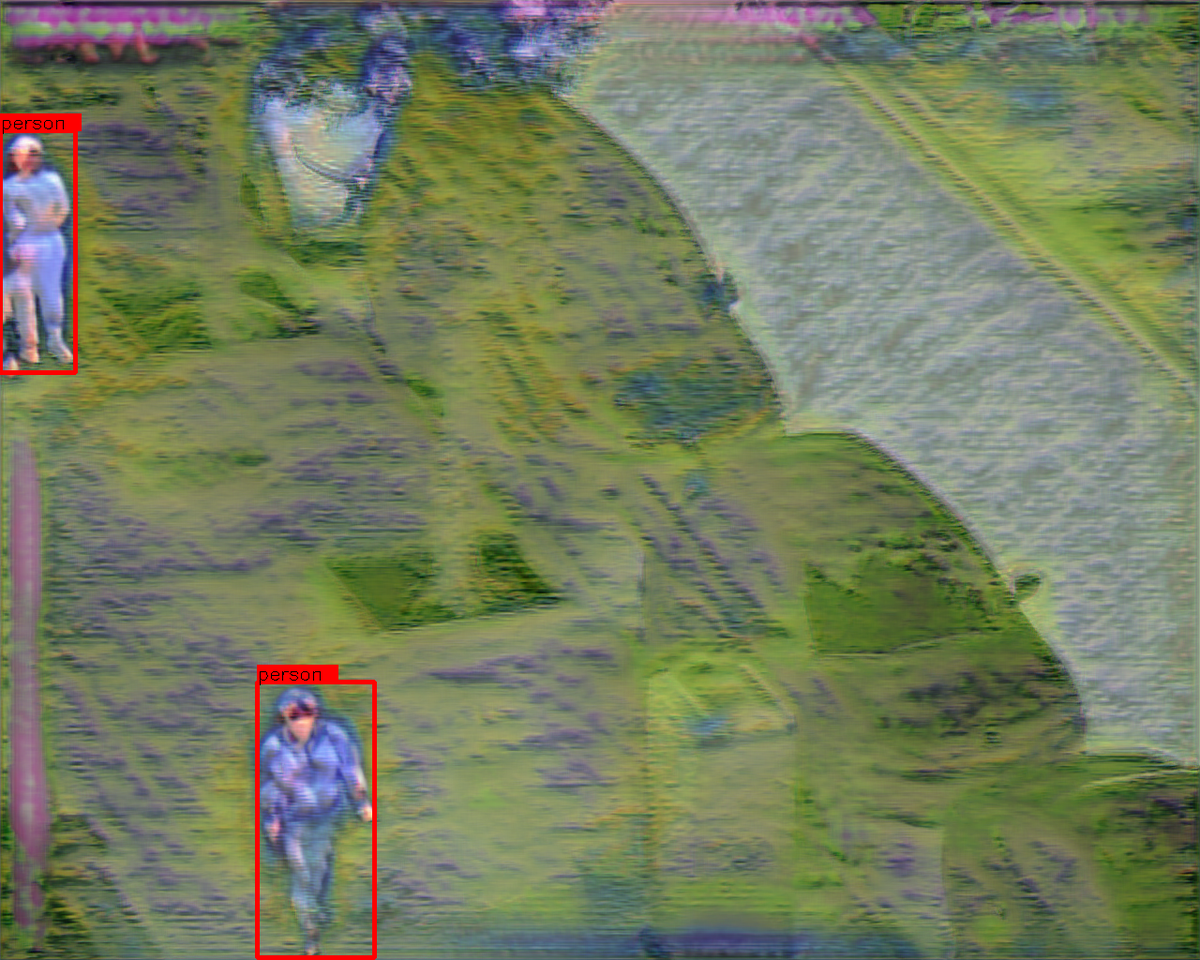}
        \includegraphics[width=\textwidth]
        {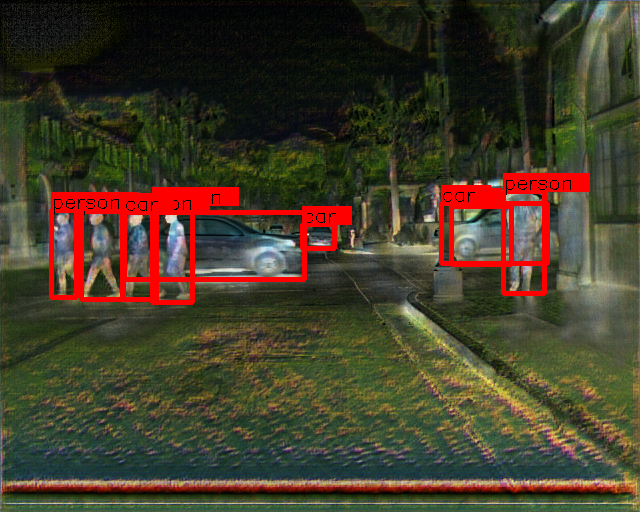}
    \end{subfigure}
    
    \\ \midrule 
    \textbf{Detector: RetinaNet} \\
    \midrule
    
    \begin{subfigure}[t]{0.23\textwidth}
        \caption{RGB - GT}
        \makebox[0pt][r]{\makebox[15pt]{\raisebox{30pt}{\rotatebox[origin=c]{90}{LLVIP}}}}%
        \includegraphics[width=\textwidth]
        {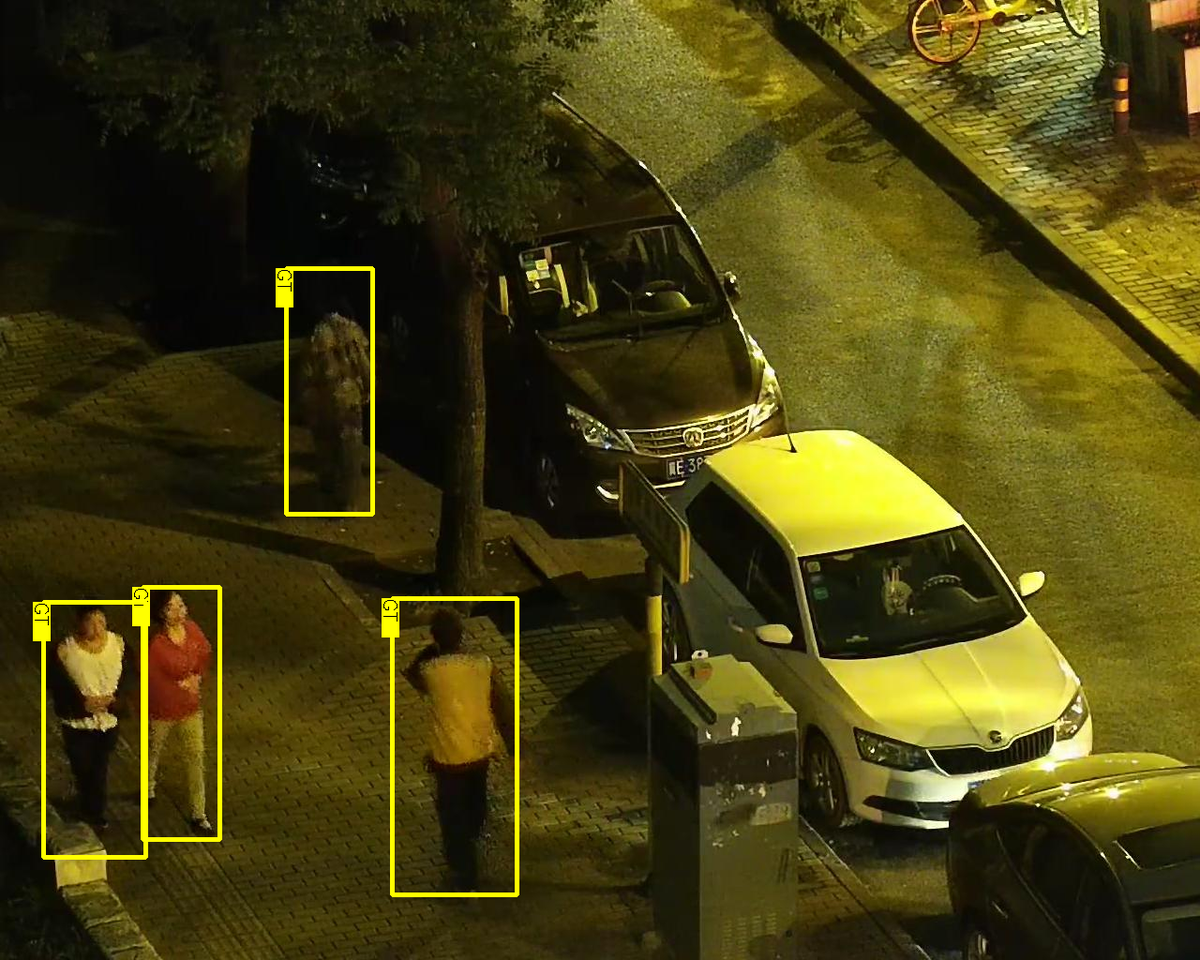}
        \makebox[0pt][r]{\makebox[15pt]{\raisebox{30pt}{\rotatebox[origin=c]{90}{FLIR}}}}%
        \includegraphics[width=\textwidth]{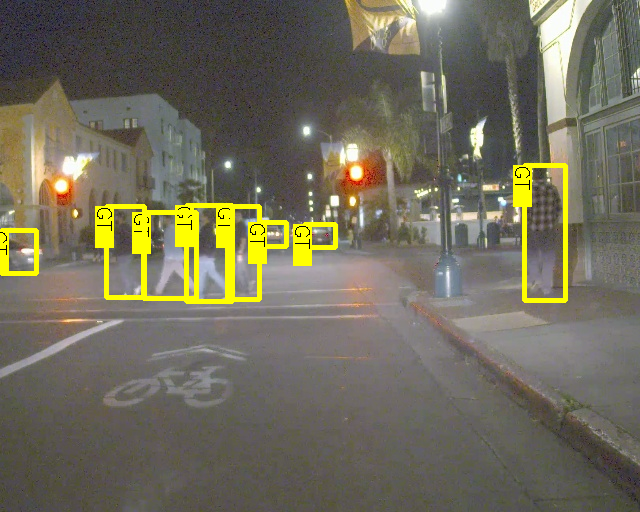}
    \end{subfigure}
    \begin{subfigure}[t]{0.23\textwidth}
        \caption{IR - FastCUT~\cite{park2020contrastive}}
        \includegraphics[width=\textwidth]{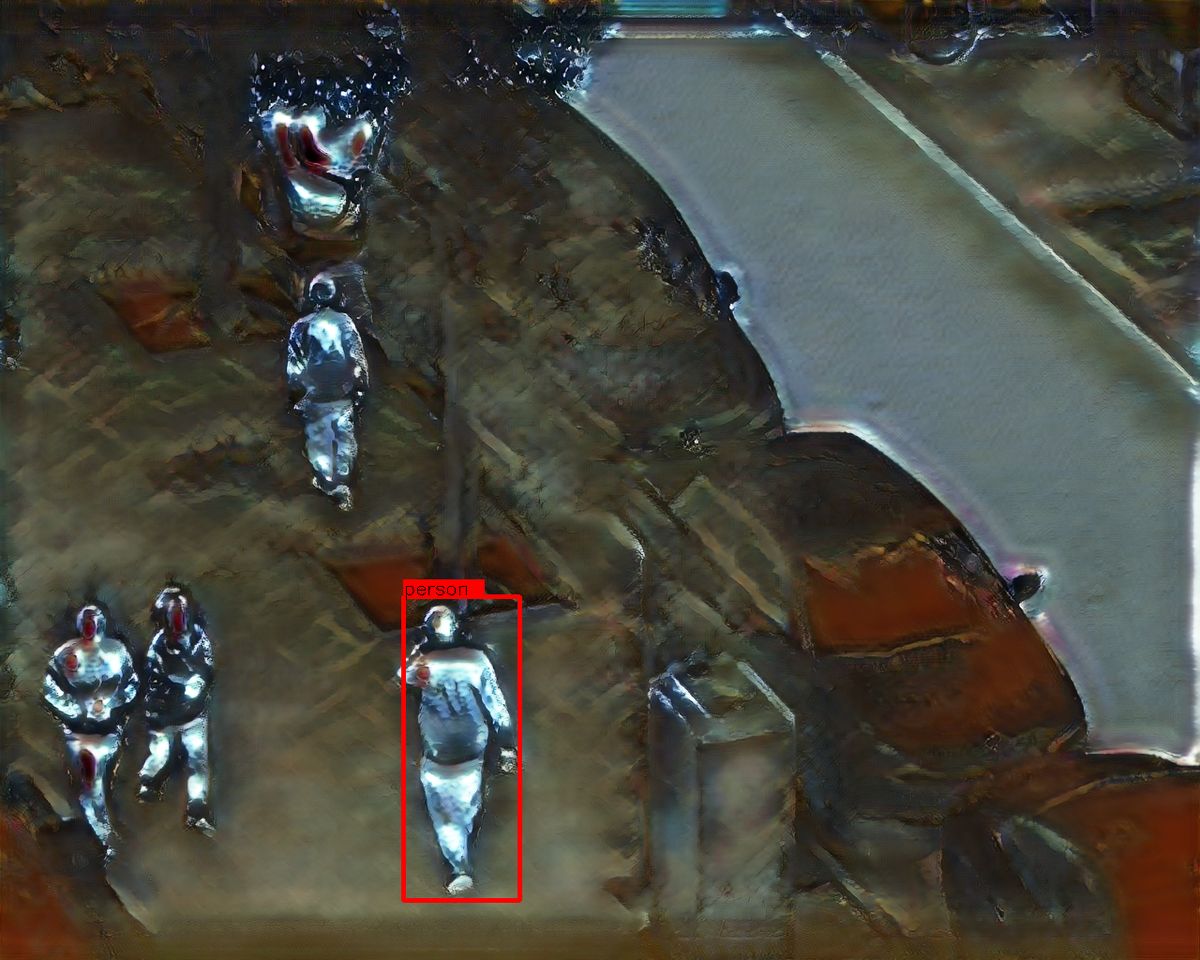}
        \includegraphics[width=\textwidth]{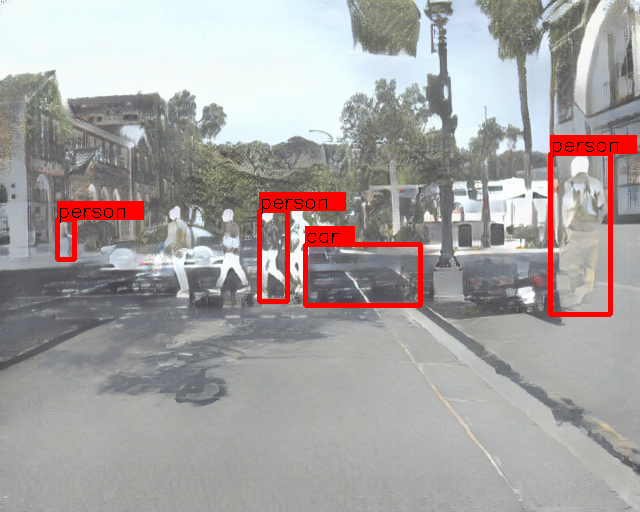}
    \end{subfigure}
    \begin{subfigure}[t]{0.23\textwidth}
        \caption{IR - Fine-tuning}
        \includegraphics[width=\textwidth]{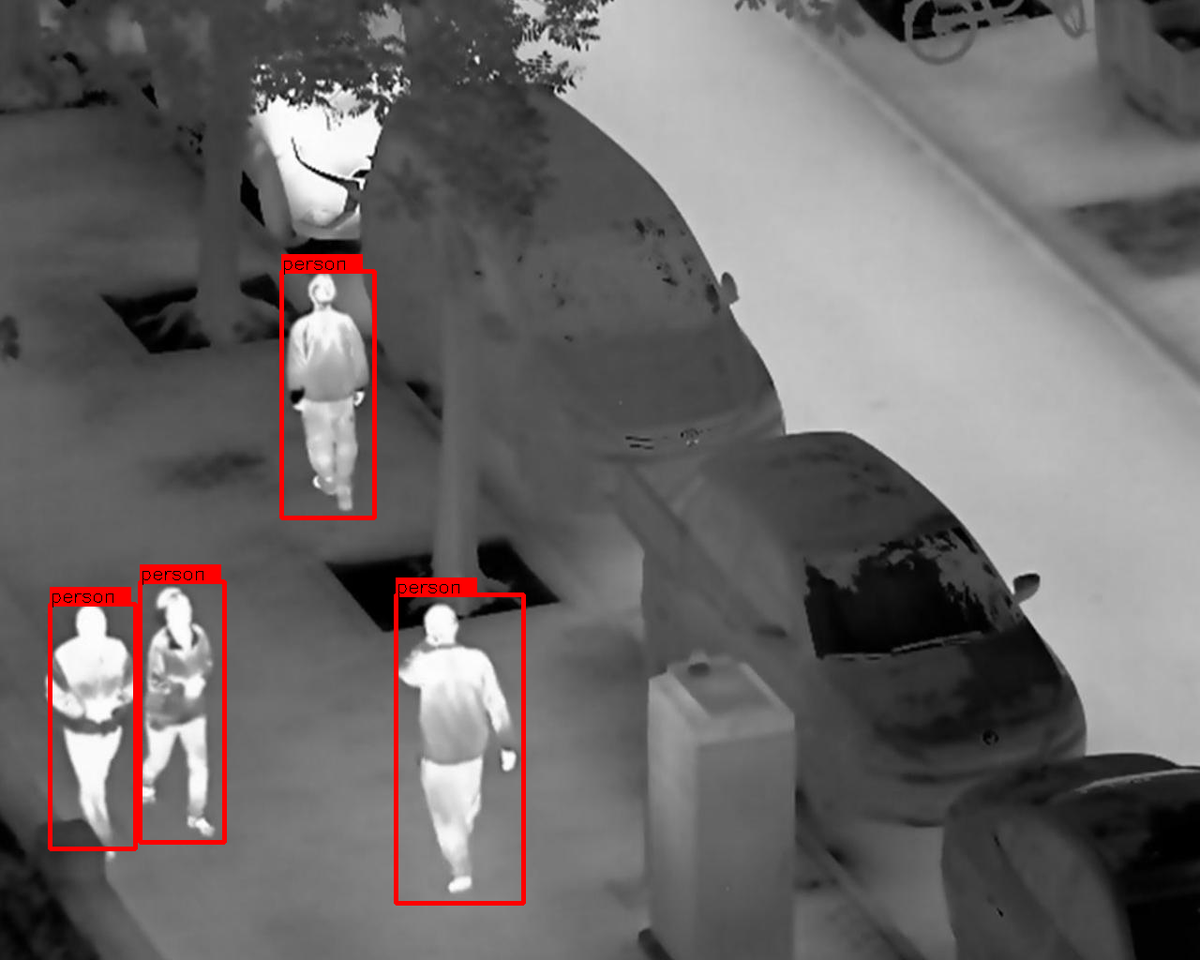}
        \includegraphics[width=\textwidth]{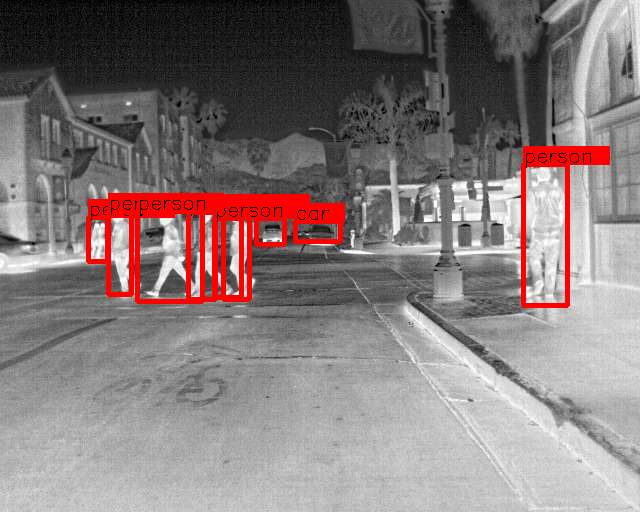}
    \end{subfigure}
    \begin{subfigure}[t]{0.23\textwidth}
        \caption{IR - ModTr (Ours)}
        \includegraphics[width=\textwidth]{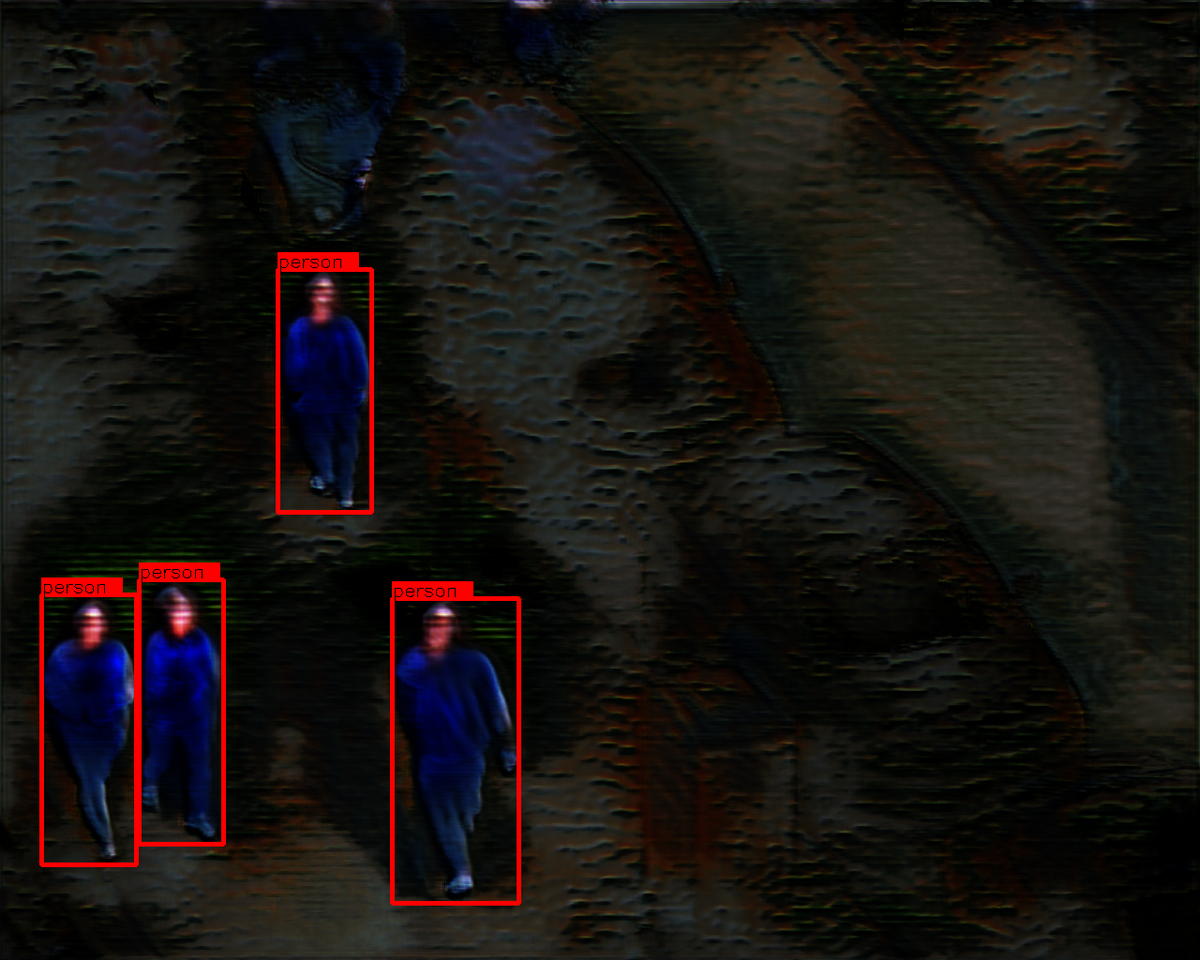}
        \includegraphics[width=\textwidth]{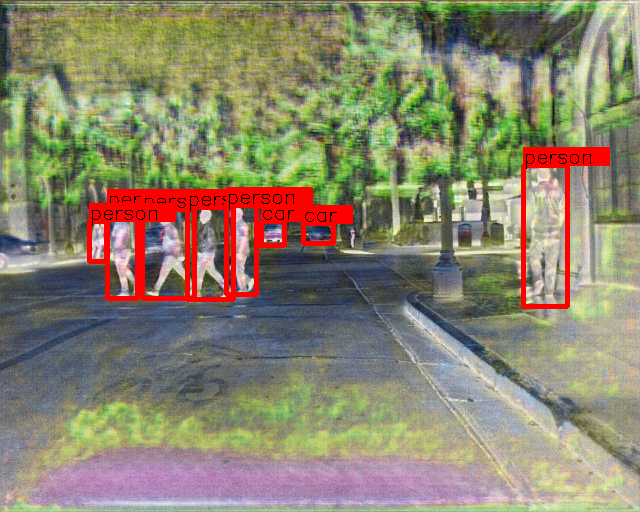}
    \end{subfigure}
    
    \\ \midrule 
    \textbf{Detector: Faster R-CNN} \\
    \midrule
    
    \begin{subfigure}[t]{0.23\textwidth}
        \caption{RGB - GT}
        \makebox[0pt][r]{\makebox[15pt]{\raisebox{30pt}{\rotatebox[origin=c]{90}{LLVIP}}}}%
        \includegraphics[width=\textwidth]{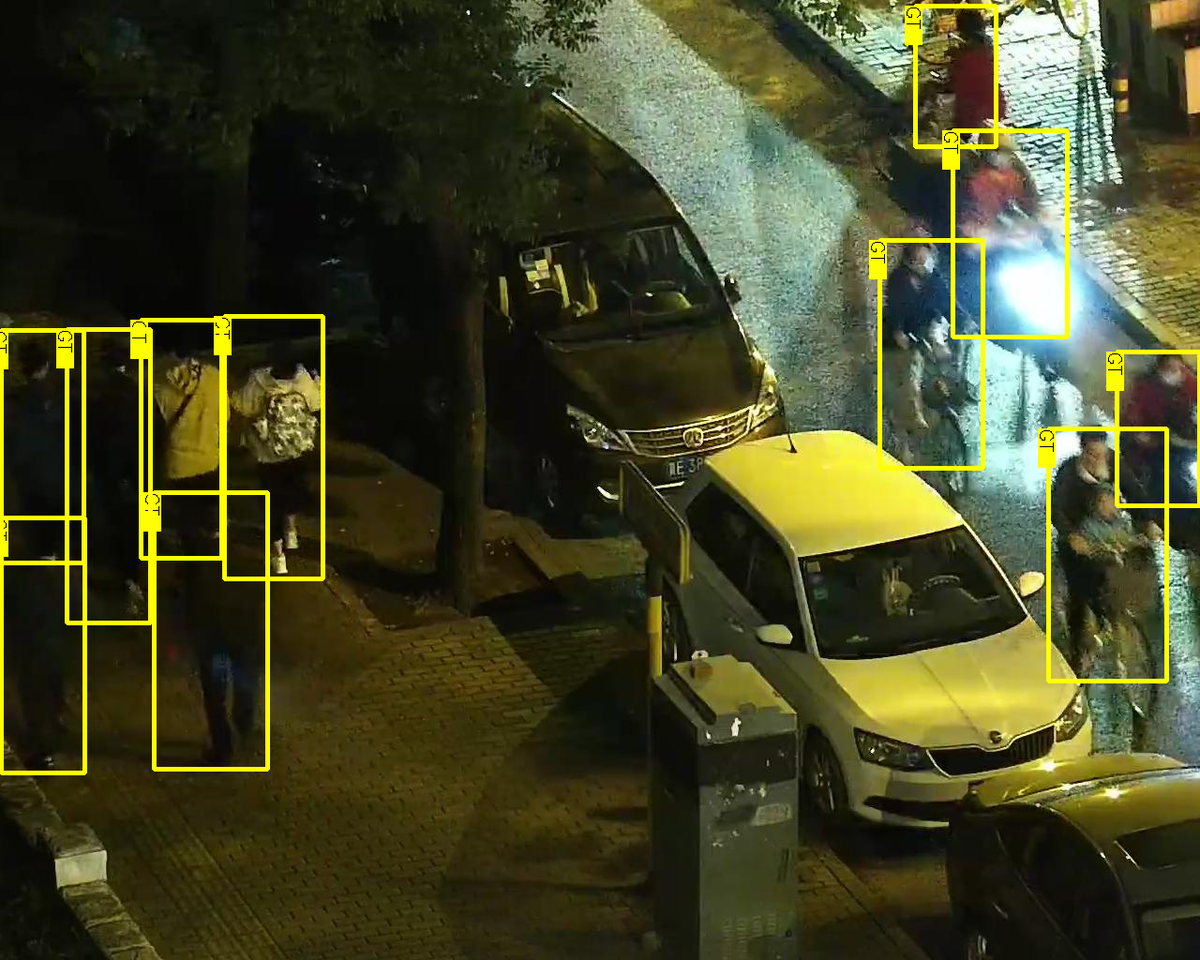}
        \makebox[0pt][r]{\makebox[15pt]{\raisebox{30pt}{\rotatebox[origin=c]{90}{FLIR}}}}%
        \includegraphics[width=\textwidth]{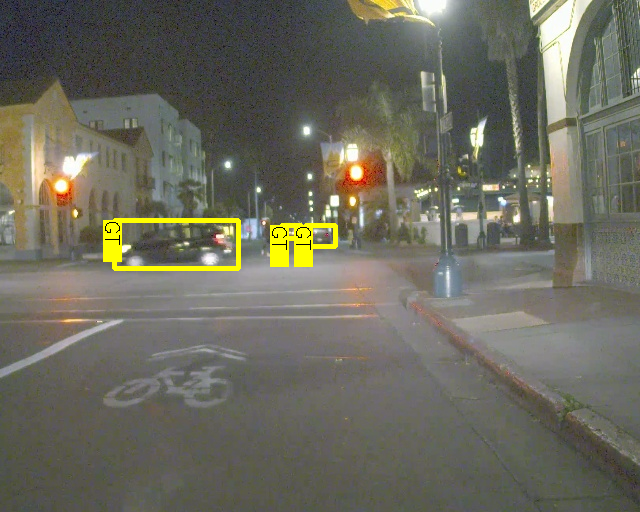}
    \end{subfigure}
    \begin{subfigure}[t]{0.23\textwidth}
        \caption{IR - FastCUT~\cite{park2020contrastive}}
        \includegraphics[width=\textwidth]{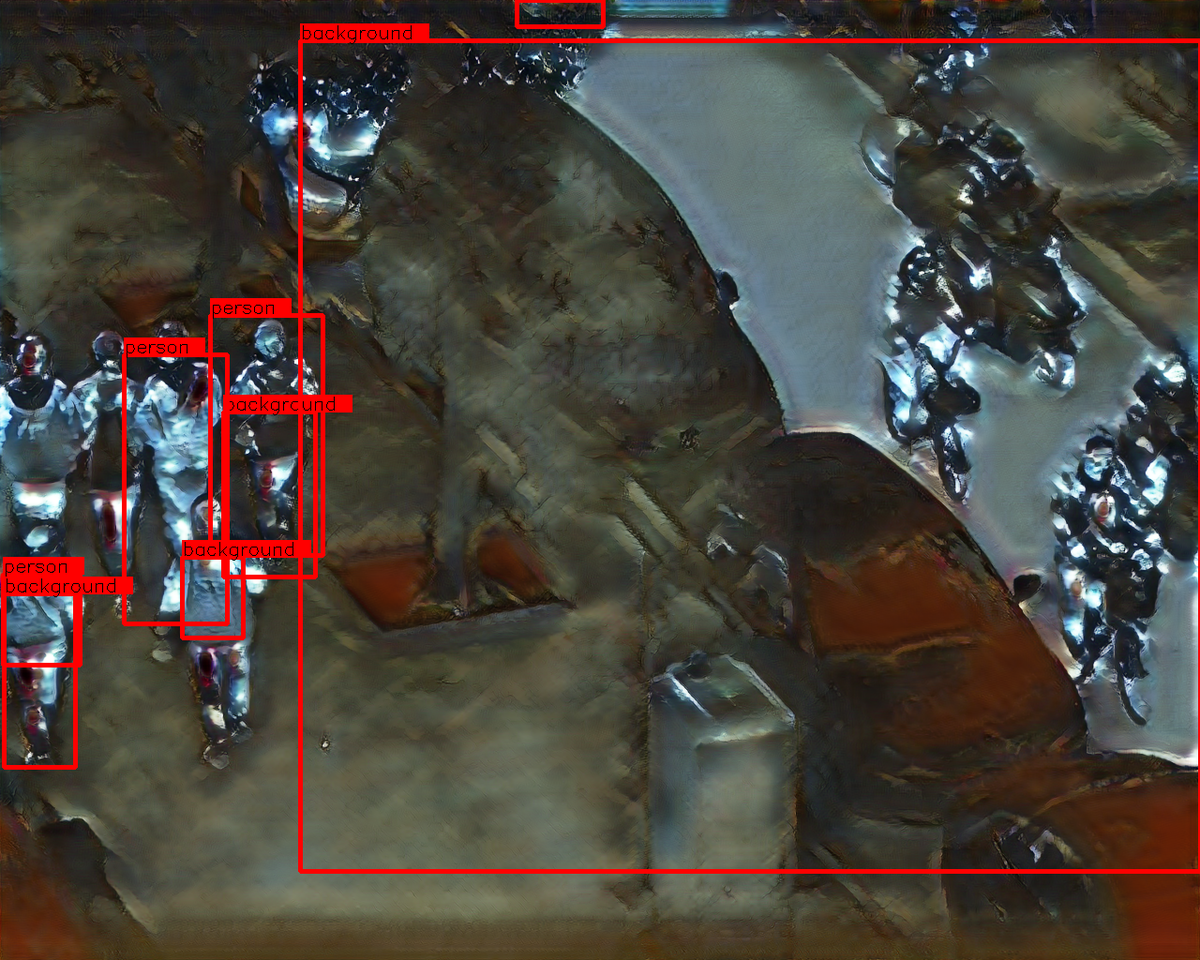}
        \includegraphics[width=\textwidth]{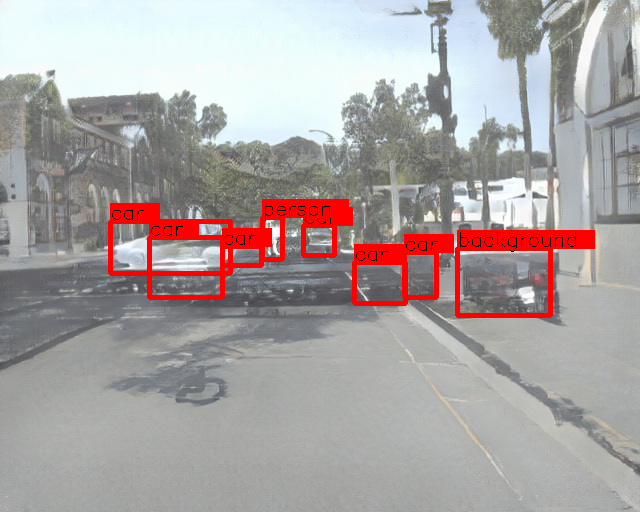}
    \end{subfigure}
    \begin{subfigure}[t]{0.23\textwidth}
        \caption{IR - Fine-tuning}
        \includegraphics[width=\textwidth]{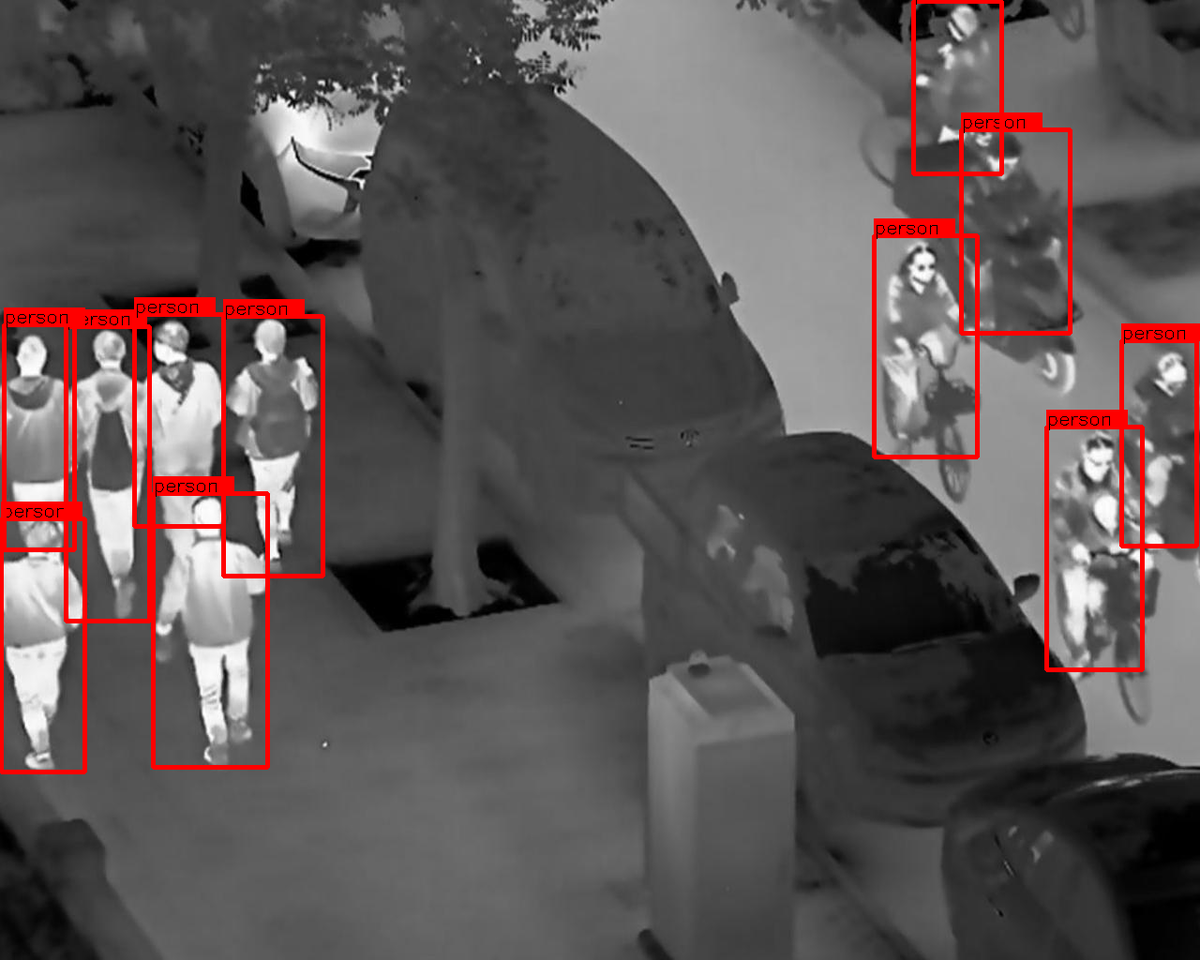}
        \includegraphics[width=\textwidth]{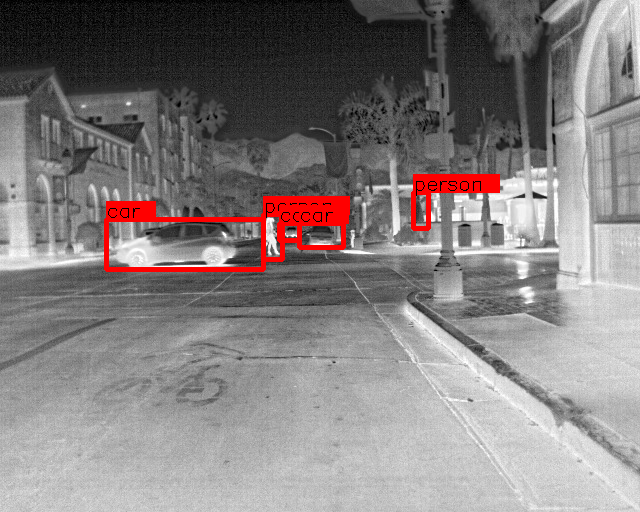}
    \end{subfigure}
    \begin{subfigure}[t]{0.23\textwidth}
        \caption{IR - ModTr (Ours)}
        \includegraphics[width=\textwidth]{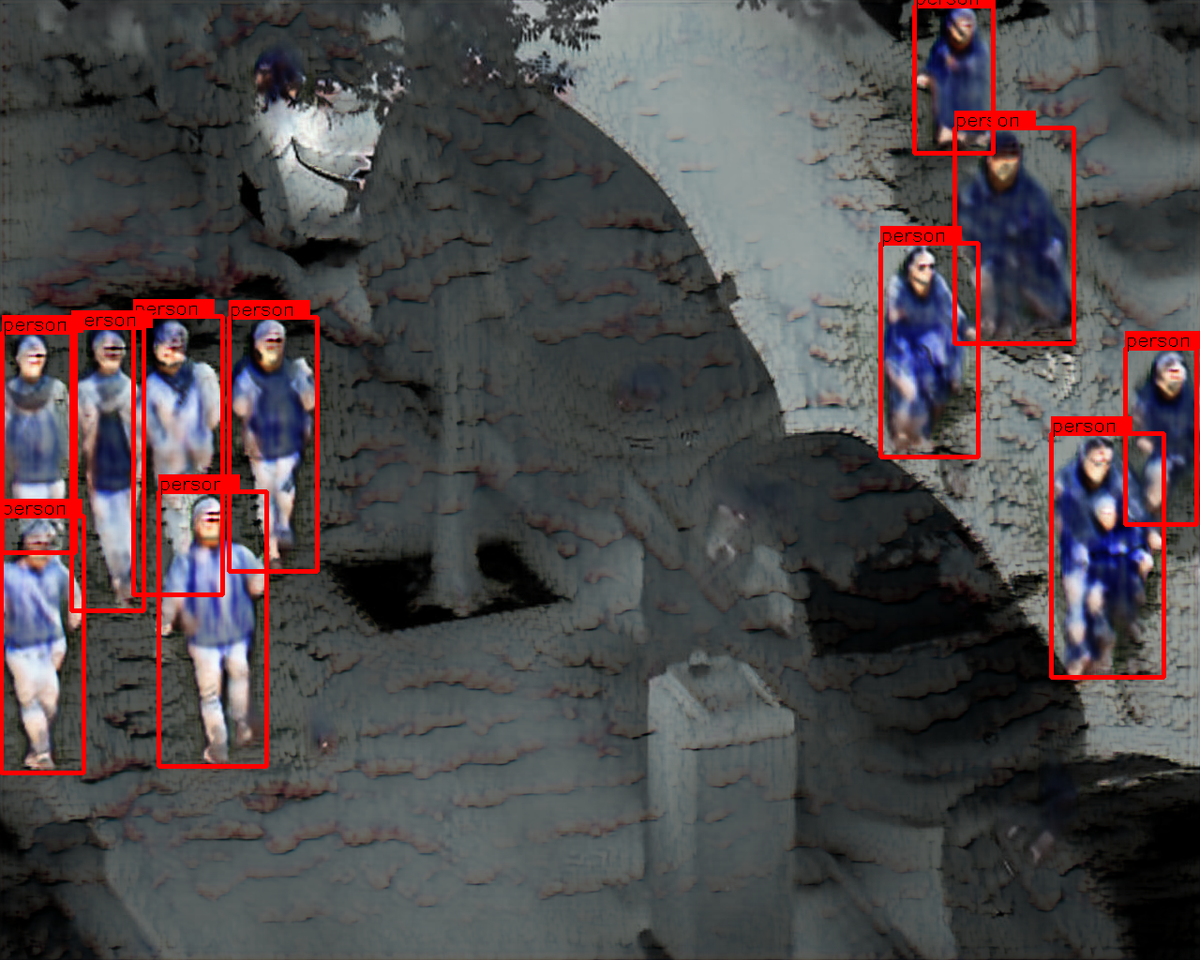}
        \includegraphics[width=\textwidth]{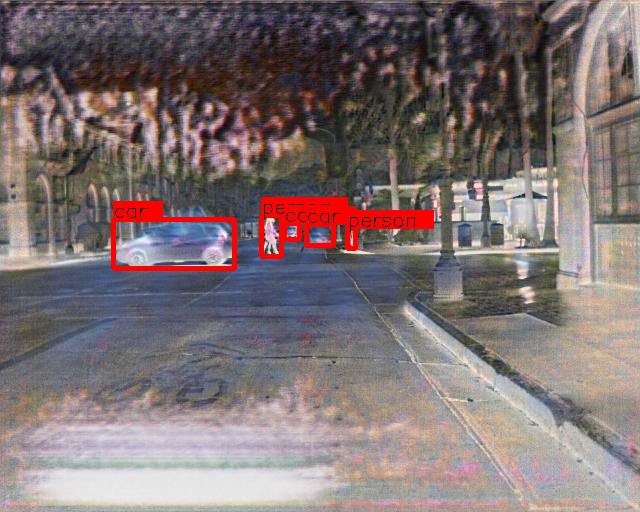}
    \end{subfigure}
    
\end{tabular}

\caption{Bounding box predictions over different OD methods for infrared images on two benchmarks: LLVIP and FLIR. Yellow and red boxes show the ground truth and predicted detections, respectively. FastCUT~\cite{park2020contrastive} is an unsupervised image translation approach that takes as input infrared images (IR) and produces pseudo-RGB images. It does not focus on detection and requires both modalities for training. Fine-tuning is the standard approach to adapting the detector to the new modality. It requires only IR data but forgets the original knowledge of the pre-trained detector. Finally, ModTr, our approach focuses on the translation on detection, requires only IR data, and does not forget the original knowledge so that it can be reused for other tasks.}
\label{fig:qualitative_results_individual}
\end{figure}

\begin{figure*}[!htp]
\centering
    \begin{tabular}{c}

    \toprule
    $\!\!\!\!$\textbf{Detector: FCOS} \\
    \midrule
    
    LLVIP Test Dataset \\
    \midrule 

    \makebox[0pt][r]{\makebox[10pt]{\raisebox{12pt}{\rotatebox[origin=c]{90}{\scriptsize GT}}}}
    \includegraphics[width=0.95\textwidth]{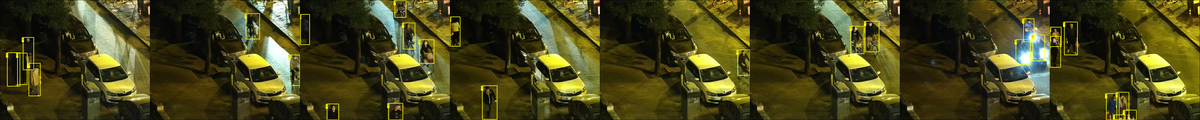} \\

    \makebox[0pt][r]{\makebox[10pt]{\raisebox{15pt}{\rotatebox[origin=c]{90}{\scriptsize FT}}}}
    \includegraphics[width=0.95\textwidth]{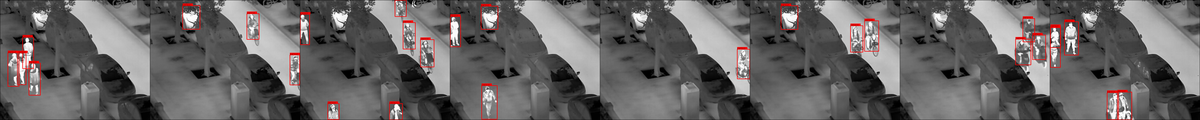} \\

    \makebox[0pt][r]{\makebox[10pt]{\raisebox{15pt}{\rotatebox[origin=c]{90}{\scriptsize FastCUT}}}}
    \includegraphics[width=0.95\textwidth]{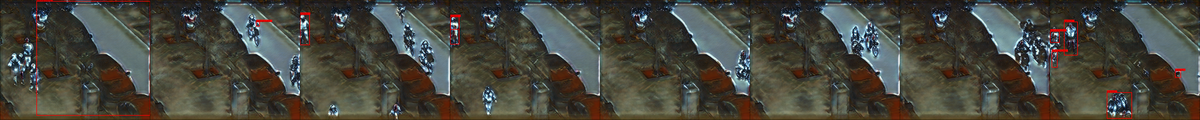} \\

    \makebox[0pt][r]{\makebox[10pt]{\raisebox{15pt}{\rotatebox[origin=c]{90}{\scriptsize $\text{ModTr}_{+}$}}}}
    \includegraphics[width=0.95\textwidth]{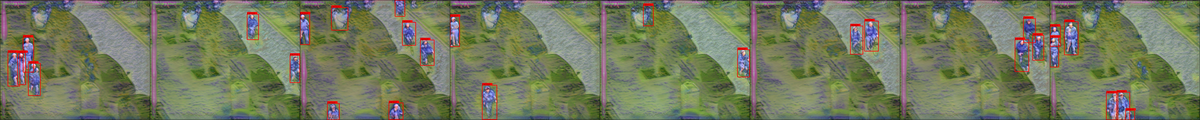} \\

    \makebox[0pt][r]{\makebox[10pt]{\raisebox{15pt}{\rotatebox[origin=c]{90}{\scriptsize $\text{ModTr}_{\odot}$}}}}
    \includegraphics[width=0.95\textwidth] {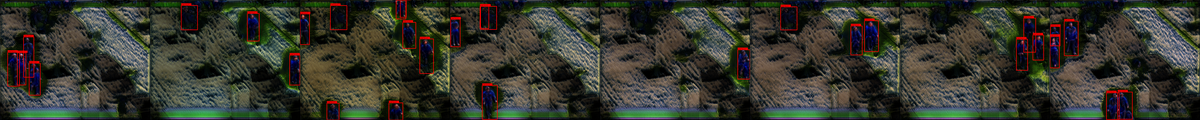} \\

    \makebox[0pt][r]{\makebox[10pt]{\raisebox{15pt}{\rotatebox[origin=c]{90}{\scriptsize $\text{ModTr}_{\oplus}$}}}}
    \includegraphics[width=0.95\textwidth] {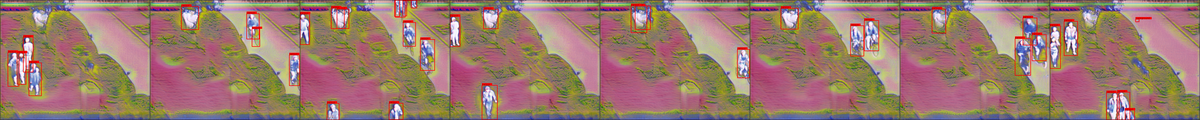} \\
    
    \midrule
    FLIR Test Dataset \\
    \midrule

    \makebox[0pt][r]{\makebox[10pt]{\raisebox{15pt}{\rotatebox[origin=c]{90}{\scriptsize GT}}}}
    \includegraphics[width=0.95\textwidth]{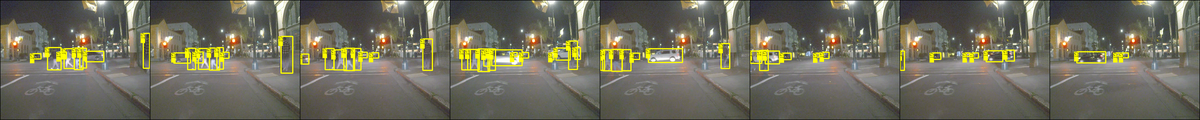} \\

    \makebox[0pt][r]{\makebox[10pt]{\raisebox{15pt}{\rotatebox[origin=c]{90}{\scriptsize FT}}}}
    \includegraphics[width=0.95\textwidth]{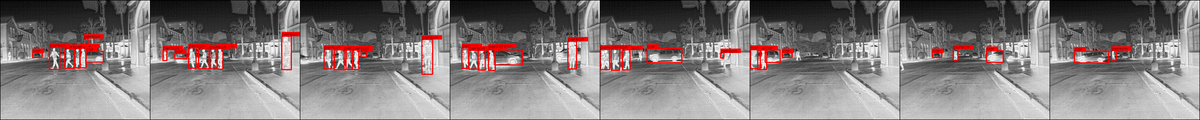} \\

    \makebox[0pt][r]{\makebox[10pt]{\raisebox{15pt}{\rotatebox[origin=c]{90}{\scriptsize FastCUT}}}}
    \includegraphics[width=0.95\textwidth]{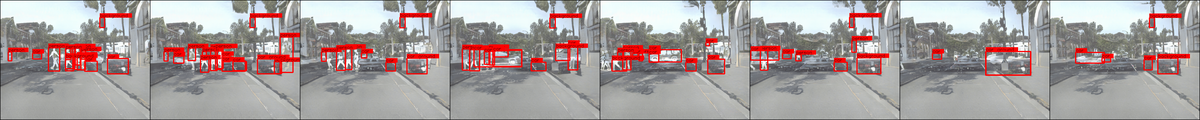} \\

    \makebox[0pt][r]{\makebox[10pt]{\raisebox{15pt}{\rotatebox[origin=c]{90}{\scriptsize $\text{ModTr}_{+}$}}}}
    \includegraphics[width=0.95\textwidth]{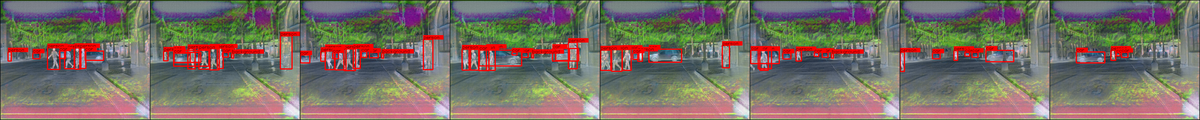} \\

    \makebox[0pt][r]{\makebox[10pt]{\raisebox{15pt}{\rotatebox[origin=c]{90}{\scriptsize $\text{ModTr}_{\odot}$}}}}
    \includegraphics[width=0.95\textwidth] {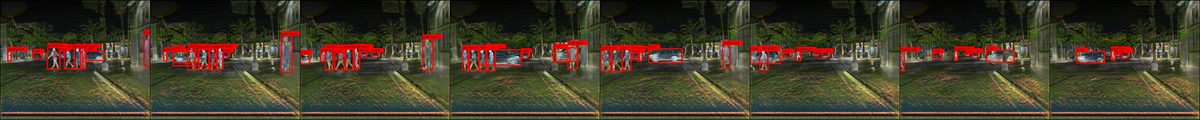} \\

    \makebox[0pt][r]{\makebox[10pt]{\raisebox{15pt}{\rotatebox[origin=c]{90}{\scriptsize $\text{ModTr}_{\oplus}$}}}}
    \includegraphics[width=0.95\textwidth] {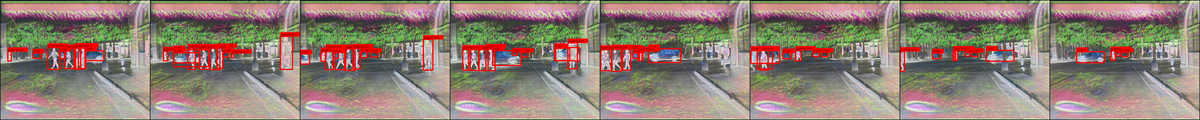} \\
    
    \bottomrule

    \end{tabular}
\caption{Illustration of a sequence of $8$ images of LLVIP and FLIR dataset for FCOS. For each dataset, the first row is the RGB modality, followed by the IR modality, followed by FastCUT~\cite{park2020contrastive}, and different representations created by ModTr and their variations.}

\label{fig:qualitative_results_fcos}

\end{figure*}

\begin{figure*}[!htp]
\centering
    \begin{tabular}{c}
    \toprule
    $\!$\textbf{Detector: RetinaNet} \\
    \midrule
    
    LLVIP Test Dataset \\
    \midrule 

    \makebox[0pt][r]{\makebox[10pt]{\raisebox{12pt}{\rotatebox[origin=c]{90}{\scriptsize GT}}}}
    \includegraphics[width=0.95\textwidth]{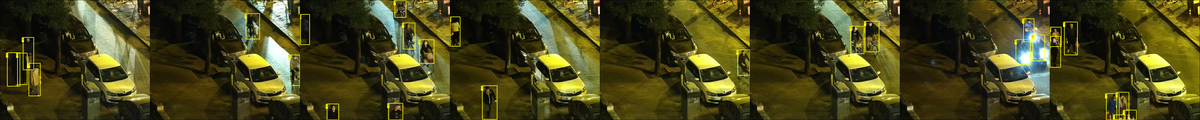} \\

    \makebox[0pt][r]{\makebox[10pt]{\raisebox{15pt}{\rotatebox[origin=c]{90}{\scriptsize FT}}}}
    \includegraphics[width=0.95\textwidth]{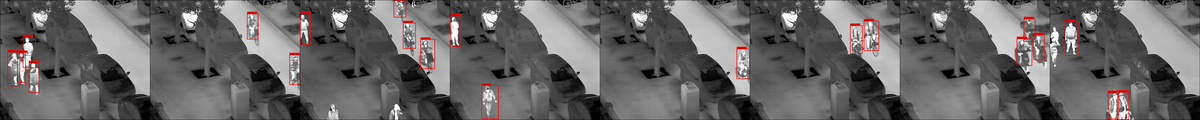} \\

     \makebox[0pt][r]{\makebox[10pt]{\raisebox{15pt}{\rotatebox[origin=c]{90}{\scriptsize FastCUT}}}}
    \includegraphics[width=0.95\textwidth]{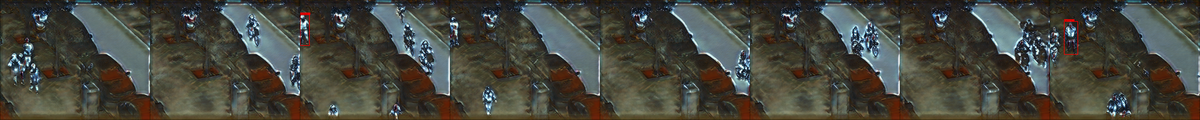} \\

    \makebox[0pt][r]{\makebox[10pt]{\raisebox{15pt}{\rotatebox[origin=c]{90}{\scriptsize $\text{ModTr}_{+}$}}}}
    \includegraphics[width=0.95\textwidth]{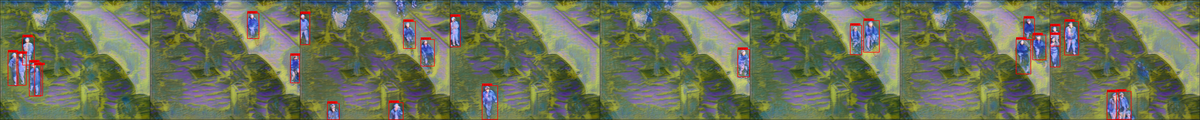} \\

    \makebox[0pt][r]{\makebox[10pt]{\raisebox{15pt}{\rotatebox[origin=c]{90}{\scriptsize $\text{ModTr}_{\odot}$}}}}
    \includegraphics[width=0.95\textwidth]{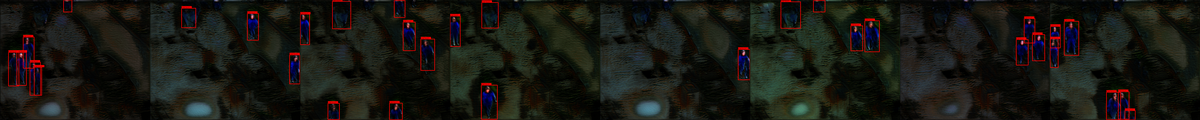}  \\

    \makebox[0pt][r]{\makebox[10pt]{\raisebox{15pt}{\rotatebox[origin=c]{90}{\scriptsize $\text{ModTr}_{\oplus}$}}}}
    \includegraphics[width=0.95\textwidth] {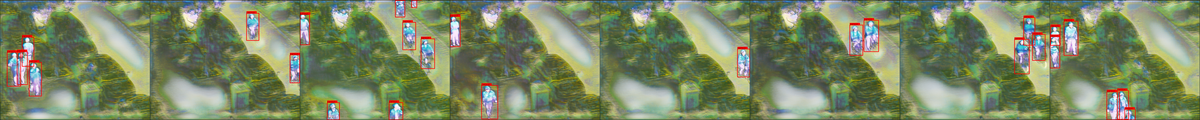} \\
    
    \midrule
    FLIR Test Dataset \\
    \midrule

    \makebox[0pt][r]{\makebox[10pt]{\raisebox{15pt}{\rotatebox[origin=c]{90}{\scriptsize GT}}}}
    \includegraphics[width=0.95\textwidth]{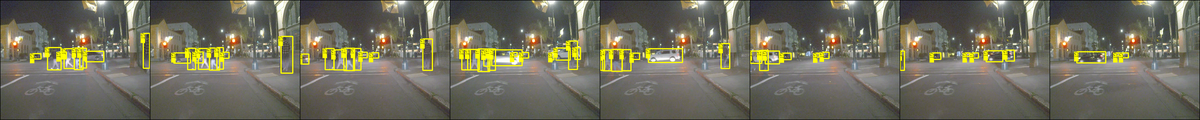} \\

    \makebox[0pt][r]{\makebox[10pt]{\raisebox{15pt}{\rotatebox[origin=c]{90}{\scriptsize FT}}}}
    \includegraphics[width=0.95\textwidth]{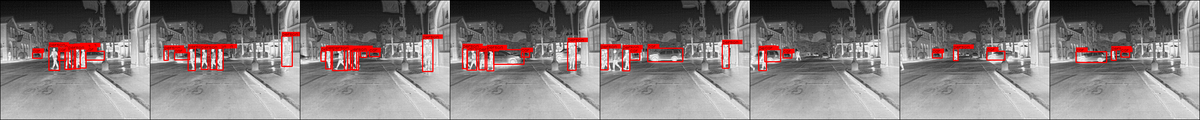} \\

    \makebox[0pt][r]{\makebox[10pt]{\raisebox{15pt}{\rotatebox[origin=c]{90}{\scriptsize FastCUT}}}}
    \includegraphics[width=0.95\textwidth]{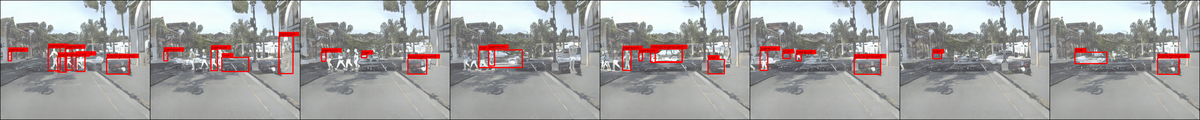} \\

    \makebox[0pt][r]{\makebox[10pt]{\raisebox{15pt}{\rotatebox[origin=c]{90}{\scriptsize $\text{ModTr}_{+}$}}}}
    \includegraphics[width=0.95\textwidth]{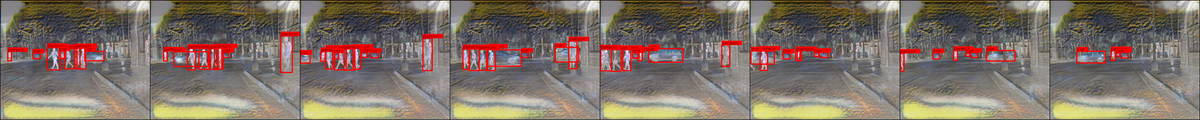} \\

    \makebox[0pt][r]{\makebox[10pt]{\raisebox{15pt}{\rotatebox[origin=c]{90}{\scriptsize $\text{ModTr}_{\odot}$}}}}
    \includegraphics[width=0.95\textwidth]{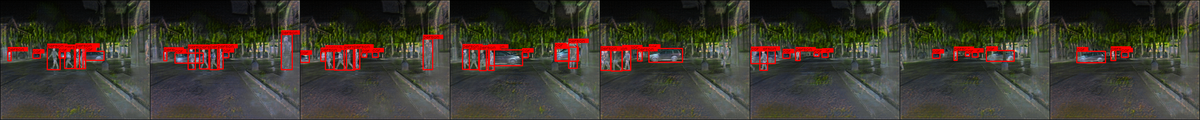} \\

    \makebox[0pt][r]{\makebox[10pt]{\raisebox{15pt}{\rotatebox[origin=c]{90}{\scriptsize $\text{ModTr}_{\oplus}$}}}}
    \includegraphics[width=0.95\textwidth] {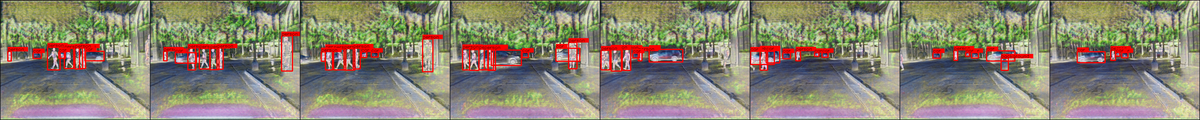} \\
    
    \bottomrule

    \end{tabular}
\caption{Illustration of a sequence of $8$ images of LLVIP and FLIR dataset for RetinaNet. For each dataset, the first row is the RGB modality, followed by the IR modality, followed by FastCUT~\cite{park2020contrastive}, and different representations created by ModTr and their variations.}

\label{fig:qualitative_results_retinanet}

\end{figure*}

\begin{figure*}[!htp]
\centering
    \begin{tabular}{c}
    \toprule
    \textbf{Detector: Faster R-CNN} \\
    \midrule
    LLVIP Test Dataset \\
    \midrule 

    \makebox[0pt][r]{\makebox[10pt]{\raisebox{12pt}{\rotatebox[origin=c]{90}{\scriptsize GT}}}}
    \includegraphics[width=0.95\textwidth]{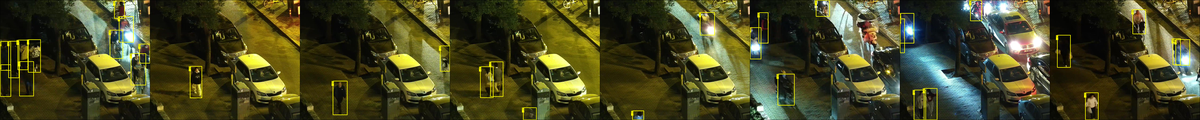} \\

    \makebox[0pt][r]{\makebox[10pt]{\raisebox{15pt}{\rotatebox[origin=c]{90}{\scriptsize FT}}}}
    \includegraphics[width=0.95\textwidth]{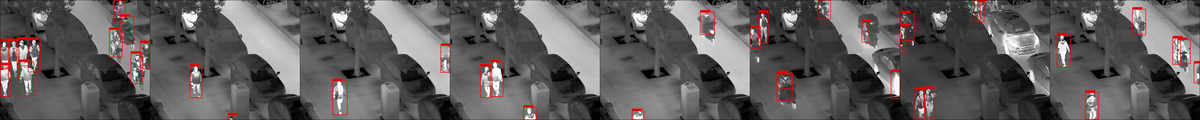} \\

     \makebox[0pt][r]{\makebox[10pt]{\raisebox{15pt}{\rotatebox[origin=c]{90}{\scriptsize FastCUT}}}}
    \includegraphics[width=0.95\textwidth]{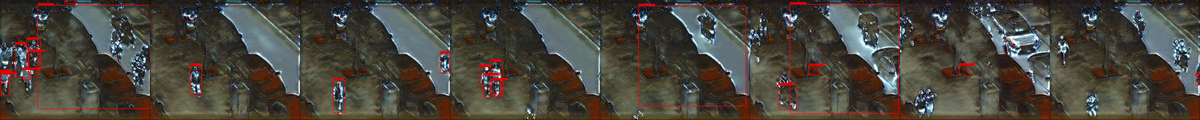} \\

    \makebox[0pt][r]{\makebox[10pt]{\raisebox{15pt}{\rotatebox[origin=c]{90}{\scriptsize $\text{ModTr}_{+}$}}}}
    \includegraphics[width=0.95\textwidth]{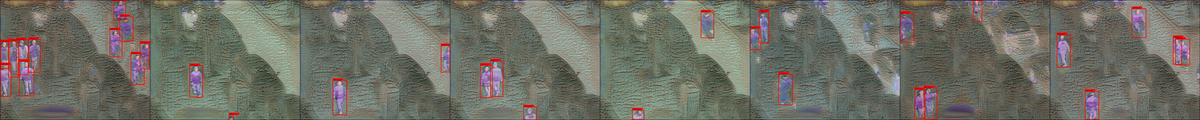} \\

    \makebox[0pt][r]{\makebox[10pt]{\raisebox{15pt}{\rotatebox[origin=c]{90}{\scriptsize $\text{ModTr}_{\odot}$}}}}
    \includegraphics[width=0.95\textwidth] {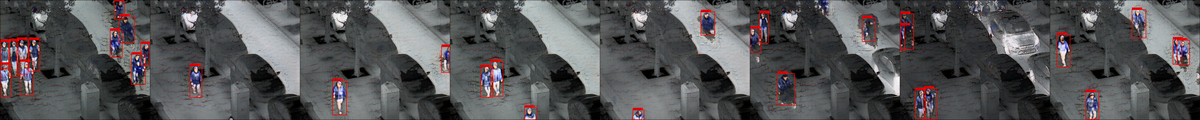} \\

    \makebox[0pt][r]{\makebox[10pt]{\raisebox{15pt}{\rotatebox[origin=c]{90}{\scriptsize $\text{ModTr}_{\oplus}$}}}}
    \includegraphics[width=0.95\textwidth]{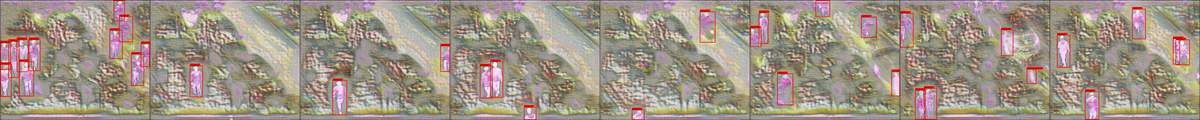} \\
    
    \midrule
    FLIR Test Dataset \\
    \midrule

    \makebox[0pt][r]{\makebox[10pt]{\raisebox{15pt}{\rotatebox[origin=c]{90}{\scriptsize GT}}}}
    \includegraphics[width=0.95\textwidth]{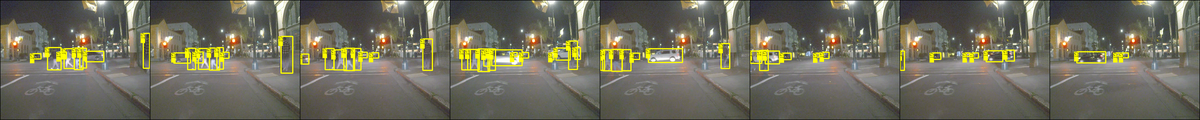} \\

    \makebox[0pt][r]{\makebox[10pt]{\raisebox{15pt}{\rotatebox[origin=c]{90}{\scriptsize FT}}}}
    \includegraphics[width=0.95\textwidth]{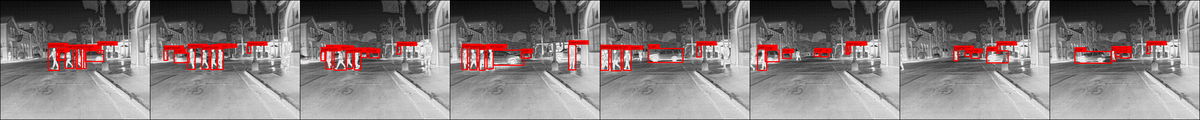} \\
    
    \makebox[0pt][r]{\makebox[10pt]{\raisebox{15pt}{\rotatebox[origin=c]{90}{\scriptsize FastCUT}}}}
    \includegraphics[width=0.95\textwidth]{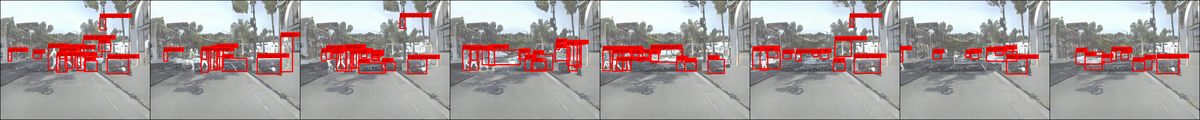} \\

    \makebox[0pt][r]{\makebox[10pt]{\raisebox{15pt}{\rotatebox[origin=c]{90}{\scriptsize $\text{ModTr}_{+}$}}}}
    \includegraphics[width=0.95\textwidth]{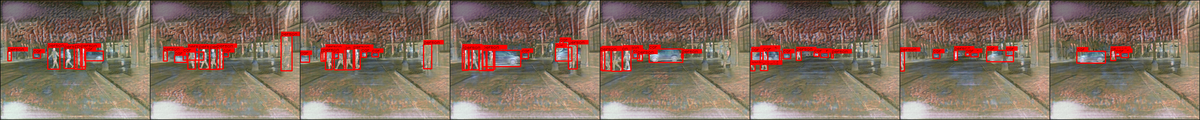} \\

    \makebox[0pt][r]{\makebox[10pt]{\raisebox{15pt}{\rotatebox[origin=c]{90}{\scriptsize $\text{ModTr}_{\odot}$}}}}
    \includegraphics[width=0.95\textwidth] {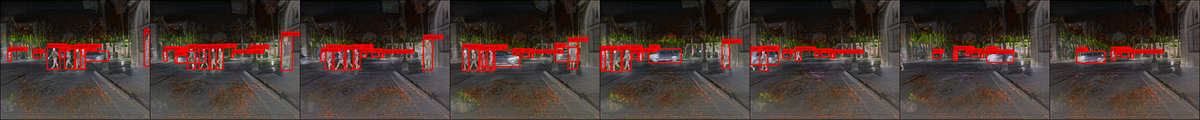} \\

    \makebox[0pt][r]{\makebox[10pt]{\raisebox{15pt}{\rotatebox[origin=c]{90}{\scriptsize $\text{ModTr}_{\oplus}$}}}}
    \includegraphics[width=0.95\textwidth]{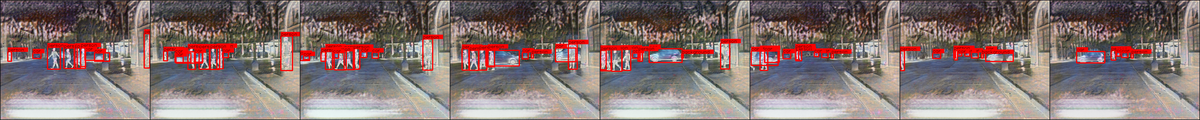} \\
    
    \bottomrule

    \end{tabular}
\caption{Illustration of a sequence of $8$ images of LLVIP and FLIR dataset for Faster R-CNN. For each dataset, the first row is the RGB modality, followed by the IR modality, followed by FastCUT~\cite{park2020contrastive}, and different representations created by ModTr and their variations.}

\label{fig:qualitative_results_fasterrcnn}

\end{figure*}

In this section, we provide additional visualizations for all the methods, including images generated by our proposal and its detection results. First, in~\fref{fig:qualitative_results_individual}, we present the detections in more detail, highlighting issues of those methods relying solely on translation, such as FastCUT, and some false positives when there is only FT. Subsequently, we provide additional visualizations for a batch of images processed by various detectors, i.e., in~\fref{fig:qualitative_results_fcos} for FCOS, in~\fref{fig:qualitative_results_retinanet} for RetinaNet, and~\fref{fig:qualitative_results_fasterrcnn} for Faster R-CNN for both datasets.

\newpage

\section{Additional Modality: Canny Edges}
\label{sec:canny_edge}

\begin{figure*}[!htp]
\centering
    \begin{tabular}{c}
    \toprule
    \textbf{Detector: Faster R-CNN} \\

    \midrule
    LLVIP Test Dataset (Canny Edge Modality) \\
    \midrule
    
    \makebox[0pt][r]{\makebox[10pt]{\raisebox{15pt}{\rotatebox[origin=c]{90}{\scriptsize FT}}}}
    \includegraphics[width=1.0\textwidth]{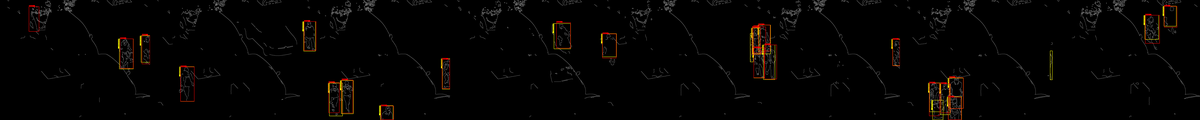} \\
    
    \makebox[0pt][r]{\makebox[10pt]{\raisebox{15pt}{\rotatebox[origin=c]{90}{\scriptsize ModTr}}}}
    \includegraphics[width=1.0\textwidth]{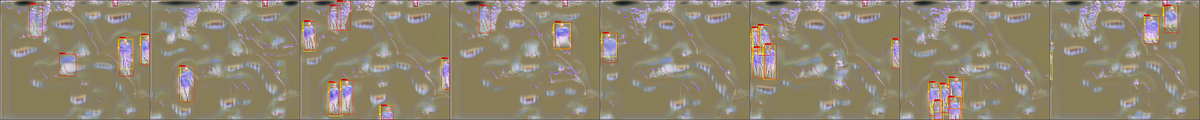} \\

    \midrule
    FLIR Test Dataset (Canny Edge Modality) \\
    \midrule

    \makebox[0pt][r]{\makebox[10pt]{\raisebox{15pt}{\rotatebox[origin=c]{90}{\scriptsize FT}}}}
    \includegraphics[width=1.0\textwidth]{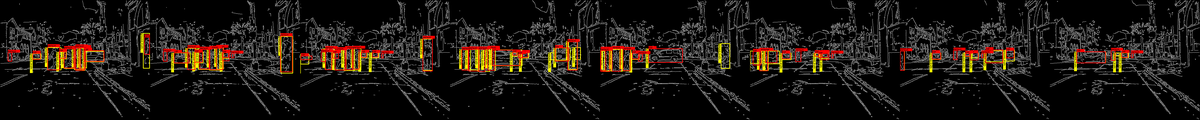} \\

    \makebox[0pt][r]{\makebox[10pt]{\raisebox{15pt}{\rotatebox[origin=c]{90}{\scriptsize ModTr}}}}
    \includegraphics[width=1.0\textwidth]{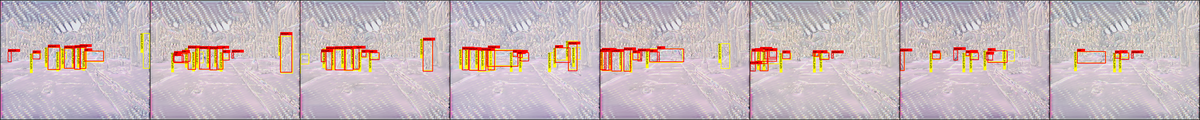} \\

    \bottomrule

    \end{tabular}
\caption{Illustration of a sequence of $8$ images from LLVIP and FLIR Canny edges test set for Faster R-CNN. The first row is the FT on the  Canny edges modality, and the second row is the result of ModTr for the LLVIP. The third row is FT on the Canny edges for FLIR, and the fourth row is the result of ModTr.}
\label{fig:canny_edge}

\end{figure*}

As described in~\cite{bachmann20244m}, edges extracted with a Canny Edge detector from IR images can be used as an additional modality. Thus, in our work, we provide some qualitative results with it as well, as illustrated by~\fref{fig:canny_edge} and quantitative results in~\tref{tab:quantitative_canny}.

\begin{table}[!htp]
\caption{Detection performance of Faster R-CNN Zero-Shot, Fine-Tuning, and ModTr on Edge Modality for LLVIP and FLIR Datasets. Edges from IR images were extracted using the Canny Edge Detector.}
\label{tab:flir_tab1}
    \centering
    \resizebox{0.8\columnwidth}{!}{%
    \begin{tabular}{lccc}

        \toprule

        \multirow{3}{*}[-1em]{\textbf{Method}} & \multicolumn{3}{c}{\textbf{Test Set Canny Edges (Dataset: LLVIP)}} \\
        
        \cmidrule(lr){2-4}
        \addlinespace[5pt]

        {}  & \multicolumn{3}{c}{\multirow{2}{*}[1em]{\textbf{Faster R-CNN}}} \\

        \cmidrule(lr){2-4}
        \addlinespace[5pt]
        
        {} &  \textbf{AP$_{50}\uparrow$} & \textbf{AP$_{75}\uparrow$} & \textbf{AP${}\uparrow$} \\

        \midrule
        
        \rowcolor[HTML]{EFEFEF}
        Zero-Shot & 14.33 ± 0.00 & 08.15 ± 0.00 & 08.01 ± 0.00 \\

        \midrule

        Fine-Tuning & 87.39 ± 1.74 & 55.50 ± 2.61 &  54.48 ± 2.40 \\

        \midrule
 
        \rowcolor[HTML]{EFEFEF}
        ModTr & 81.23 ± 1.53 & 48.66 ± 1.57 & 47.06 ± 1.18 \\

        \midrule
            \multirow{3}{*}[-1em]{\textbf{Method}} & \multicolumn{3}{c}{\textbf{Test Set Canny Edges (Dataset: FLIR)}} \\
        
        \cmidrule(lr){2-4}
        \addlinespace[5pt]

        {}  & \multicolumn{3}{c}{\multirow{2}{*}[1em]{\textbf{Faster R-CNN}}} \\

        \cmidrule(lr){2-4}
        \addlinespace[5pt]
        
        {} &  \textbf{AP$_{50}\uparrow$} & \textbf{AP$_{75}\uparrow$} & \textbf{AP${}\uparrow$} \\

        \midrule
        
        \rowcolor[HTML]{EFEFEF}
        Zero-Shot & 18.43 ± 0.00 & 06.68 ± 0.00 & 08.44 ± 0.00 \\

        \midrule

        Fine-Tuning & 60.43 ± 2.05 & 26.38 ± 0.28 & 30.88 ± 0.88 \\

        \midrule
 
        \rowcolor[HTML]{EFEFEF}
        ModTr & 51.81 ± 1.41 & 22.14 ± 0.83 & 25.59 ± 0.83 \\

        \bottomrule
    \end{tabular}
    }
\label{tab:quantitative_canny}
\end{table}

\noindent Here, we clarify that our approach is dependent on the zero-shot capability of the model to incorporate knowledge of the translation network. So, we can incorporate other modalities as well as the IR, but if the new modality does not have a good zero-shot like the Canny Edges, our method can incorporate the new knowledge but not surpass the fully fine-tuning (FT). But it is also important to mention that even without surpassing for edges, we were able to keep prior knowledge and reach good performance even though we still have a gap for the edges version when compared with FT.

\bibliographystyle{splncs04}
\bibliography{egbib}
\end{document}